\pdfoutput=1
\documentclass[10pt,twocolumn,letterpaper]{article}
\newif\ifarxiv
\newif\ifalgorithmshown
\algorithmshownfalse   %
\arxivtrue   %
\ifarxiv\usepackage[pagenumbers]{cvpr}\else\usepackage[review]{cvpr}\fi

\usepackage{cuted} %

\usepackage{currfile} %

\usepackage{caption} %

\usepackage{newtxtext}   %
\usepackage{booktabs}
\expandafter\def\csname ver@fixltx2e.sty\endcsname{}   %
\usepackage{dblfloatfix}   %
\usepackage{multicol}   %
\usepackage{array}
\usepackage{algorithm}
\usepackage[noend]{algpseudocode}
\usepackage{amsmath,amssymb}
\usepackage{dsfont}   %
\usepackage{microtype}
\usepackage{xcolor}
\usepackage{tikz}
\usetikzlibrary{arrows.meta,positioning,fit,backgrounds,calc}
\definecolor{cvprblue}{rgb}{0.21,0.49,0.74}
\usepackage[pagebackref,breaklinks,colorlinks,allcolors=cvprblue]{hyperref}
\def\paperID{*****}
\def\confName{CVPR}
\def\confYear{2027}
\newcommand{\gspool}{\textsc{GS-Pool}}
\newcommand{\gsdiff}{\textsc{GS-Diff}}
\newcommand{\oscd}{\textsc{O-SCD}}
\newcommand{\mvcd}{\textsc{MV-3DCD}}
\newcommand{\Gone}{\mathcal{G}_{T_1}}
\newcommand{\Gtwo}{\mathcal{G}_{T_2}}

\title{\gspool: Object-Level Change Detection in 3D Gaussian Splatting}
\author{Boaz Keren-Gil \qquad James Gain \qquad Patrick Marais\\
University of Cape Town}

\begin{document}
\maketitle
\begin{strip}
\centering
\includegraphics[width=\textwidth]{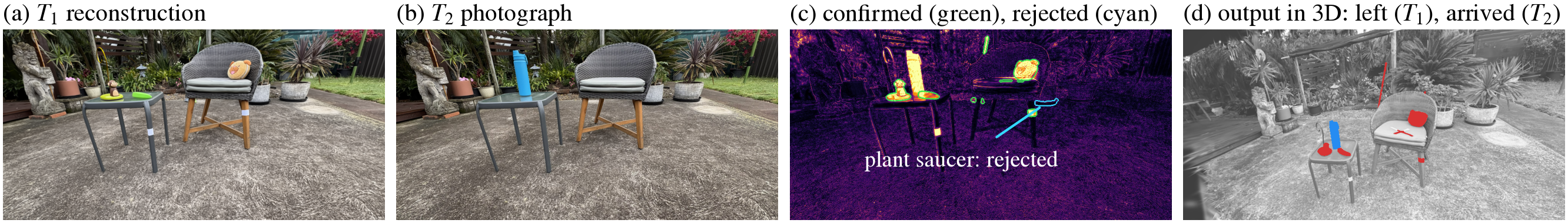}
\captionsetup{hypcap=false}
\captionof{figure}{\gspool{} on Garden. The photographs confirm or reject each proposed object (c), and each visit returns what changed (d).}
\captionsetup{hypcap=true}
\label{fig:teaser}
\end{strip}

\begin{abstract}
Factories, museums and surveyors photograph the same space months apart and need to know which objects changed. When each visit is reconstructed with 3D Gaussian Splatting (3DGS)~\cite{3dgs}, a direct comparison of the two reconstructions does not answer this. Training is stochastic, so two reconstructions of an unchanged space never coincide, and the second visit is often a quick re-scan with far fewer photographs. We propose \gspool{}, which takes two independently reconstructed Gaussian fields of the same space and returns the changed objects in each, together with their masks. SAM2 masks of each visit's photographs are lifted onto the Gaussians that render them and merged into an \emph{object pool}, so every decision is taken once per object in 3D. We introduce a \emph{photographic carrier}, the 3DGS training loss of each input reconstruction against the other visit's photographs, backpropagated to the Gaussians that rendered each pixel. We combine it with \gsdiff's geometry and colour terms and our distilled DINOv3 features. This evidence is compared with that of the objects present in both visits, which sets a change threshold for each scene. On PASLCD~\cite{paslcd}, \gspool{} reaches mIoU/F1 scores of 0.751/0.846 against 0.644/0.758 for \gsdiff{}, the strongest prior method, a gain of 17\%/12\%. Its mIoU is also 36\%, 40\% and 57\% above that of \oscd{}, PlenoCI~\cite{plenoci} and \mvcd{}, and it reaches 0.855 mIoU on CL-Splats, 33\% above \mvcd{}. Each changed object is returned as a set of Gaussians with the evidence behind its decision, which an inspector can review in 3D.
\end{abstract}

\section{Introduction}

Consider a room photographed on two visits, several months apart. Between the
visits, objects in this room have changed, the
light is different and the camera path is different. We assume that the two
reconstructions of the room are built independently with 3D Gaussian Splatting (3DGS). They share no
primitives, and their primitives differ in number and placement. The goal is to report only which objects changed, a question many factories, museums and surveyors
ask.

Change detection in 3D Gaussian Splatting is usually posed as a
\emph{dense} problem. For example, \mvcd{} and \oscd{} fuse per-view cues into a change
field over the Gaussians. \gsdiff{}~\cite{gsdiff}, the strongest prior
method, compares the two primitive sets directly, renders a change score per primitive and thresholds the rendered score at a
fixed $0.5$ in every pixel.
These methods make pixel-level decisions, with a threshold fixed in advance, about objects that may span thousands of primitives.

Two reconstructions of the same
unchanged space never coincide, because 3DGS
training is stochastic and each visit is photographed along a different camera
path. We refer to this difference as \emph{drift}, as \gsdiff{}~\cite{gsdiff} does. A difference computed between
the two reconstructions therefore mixes change with drift. Unlike the primitives, the photographs that were used for training
carry no reconstruction drift. \gspool{}
therefore lets the reconstructions propose a set of candidate object changes directly from the primitive space,
and the photographs of the other visit confirm or reject them. Geometry and colour alone can miss a different object in the same place, so we also distill a \emph{semantic field}, generated from a foundation model, into the
primitives.
\gspool{} keeps \gsdiff's drift model, geometry and colour
kernels and observability weight unchanged. We contribute:
\begin{itemize}\setlength\itemsep{1pt}
\item an \emph{object pool}: SAM2 masks of each visit's photographs, assigned
to the primitives that render them (\emph{lifted}) and merged across views into
objects, so evidence is aggregated and decided once per object and not per primitive
(Appendix~\ref{sec:poolrecord}).
\item a \emph{photographic carrier}: the 3DGS training loss of each input
reconstruction against
the \emph{other} visit's photographs, backpropagated
through the rasteriser
to the primitives that rendered each pixel
(Figure~\ref{fig:photo}).
\item a change threshold set on each scene: each object's evidence is compared
with that of the objects present in both visits, which form an empirical
\emph{null} distribution of unchanged objects.
\item an evaluation against \mvcd{} and \oscd{} on identical inputs, with three
reconstructions of each visit from each of two trainers, false alarms measured on relit
captures of unchanged scenes, and change annotations for the five real scenes of
CL-Splats~\cite{ackermann2025clsplats} in PASLCD's convention
(Appendix~\ref{sec:clsplats}).
\end{itemize}

\section{Related work}

\textbf{Change detection in image space.} The longest line of prior work compares two
photographs of the same place. Early methods compute or learn a pixel- or superpixel-wise
difference between paired viewpoints~\cite{sakurada_change_2015,
alcantarilla2018street, daudt2018fully, varghese2018changenet}. Later methods also accommodate viewpoint change with
supervision~\cite{cyws2d, cyws3d, lin2025robust}. The most recent methods drop
change labels and build on frozen foundation models~\cite{kim2025towards,
cho2025zero, goldilocs}. For all these methods, the unit of decision is either a
pixel or a bounding box in one image pair. In terms of change detection
performance, the strongest 2D method, CYWS-2D, evaluated in~\cite{mv3cd},
reaches 0.273 mIoU on PASLCD.

\textbf{Change detection with 3D reconstructions.} Rendering one visit's
reconstruction using the other visit's camera pose removes the need for matching
viewpoints~\cite{kruse2024splatpose, liu2024splatposeplus, zhou2023pad,
lu20253dgs, jiang2025gaussian}. Recent methods evaluated on PASLCD decide
per primitive. \mvcd{} thresholds per-view 2D change maps and fuses them into a change
channel optimised on the Gaussians~\cite{mv3cd}. \oscd{}~\cite{oscd}
localises each incoming image against the reference splat and fuses per-view
cues by a self-supervised objective. Its per-view cue is the 3DGS training loss against each
incoming frame, fitted into a separate per-Gaussian change parameter. Our
photographic carrier instead carries that loss through the rasteriser onto each
visit's own primitives, in both directions between two independent
reconstructions, and pools it per object. \gsdiff{}~\cite{gsdiff} compares the two primitive sets
directly, modelling reconstruction drift and weighting each primitive by its
Fisher information. \gsdiff{} is the closest method to ours, and its kernels
are adopted unchanged (Section~\ref{sec:evidence}). Parallel work,
PlenoCI~\cite{plenoci}, compares renders of two instance-aware reconstructions
along image edges, with thresholds selected per scene, and classifies each change
as geometric or appearance-based through analytic plenoptic derivatives. PlenoCI works well on unchanged scenes, flagging few pixels between two of their reconstructions, but on PASLCD's changed scenes its mIoU is below that of state-of-the-art methods. \gspool{} instead runs on standard reconstructions and calibrates its change
threshold on each scene. Galappaththige et al.~\cite{u3dgs} attribute reconstruction residuals to
primitives as an uncertainty measure that suppresses false detections, while our
photographic carrier instead attributes the training loss against the other visit's
photographs, as evidence of change.

\textbf{Object-level change detection.} These methods typically make per-pixel or
per-primitive decisions. SceneDiff~\cite{scenediff} decides per object, similar to us. It introduces a benchmark of
paired video clips with consistent object identities across visits, and it compares foundation model features between overlapping frames, with
depth and poses predicted by a feed-forward network. It builds no
reconstruction, and its output is a mask per frame. \gspool{} takes two
independently trained reconstructions and returns each changed object as a set
of primitives in its reconstruction. Thus, while both methods decide per object, their inputs and outputs differ.

\textbf{Model-to-model comparison.} Comparing two models of a place predates
radiance fields, in bitemporal point clouds~\cite{c2c, LAGUE201310,
yew2021city, de2023siamese}, robotic maps~\cite{schmid2022panoptic, fu2022planesdf, zhu2024living} and between photographs and
a city model~\cite{taneja2011image}. For splats the question
has mostly been posed as update~\cite{cheng2025lt, zeng2025gaussianupdate,
yugay2025gaussian, ackermann2025clsplats} or as registration assuming no
change~\cite{gaussreg, siren}. Most point-cloud methods (SLAM-based) compare geometry alone, and
map-based methods maintain one map that is updated as a robot revisits a place.
The splat update methods localise change in order to retrain the changed region
of an existing reconstruction from new photographs, rather than comparing two
independent reconstructions.

\textbf{Semantic fields.} Image foundation model features are often distilled
into neural radiance fields and Gaussian splats~\cite{dff, n3f, lerf, feat3dgs, langsplat,
gaussiangrouping}, where the added feature channels support editing,
querying or segmentation of a scene. We distill DINOv3~\cite{dinov3} features into each reconstruction with its Gaussians held fixed, and use the distilled
features as one source of change evidence. By contrast, SAM2~\cite{sam2} only segments the
photographs into object masks (Section~\ref{sec:pool}).

\textbf{Benchmarks.} PASLCD~\cite{paslcd, mv3cd} has five indoor
and five outdoor scenes, each in two instances, one under consistent and one under varied lighting. Each
instance pairs a reference capture of 50 to 109 photographs with an inference
capture of 25 annotated test views, and a joint COLMAP solve. CL-Splats~\cite{ackermann2025clsplats}
photographed five real scenes twice, with objects changed between the two captures. No change masks are available, so we annotated the masks ourselves (Appendix~\ref{sec:clsplats}).

\section{Method}

\begin{figure*}[t]
\centering
\input{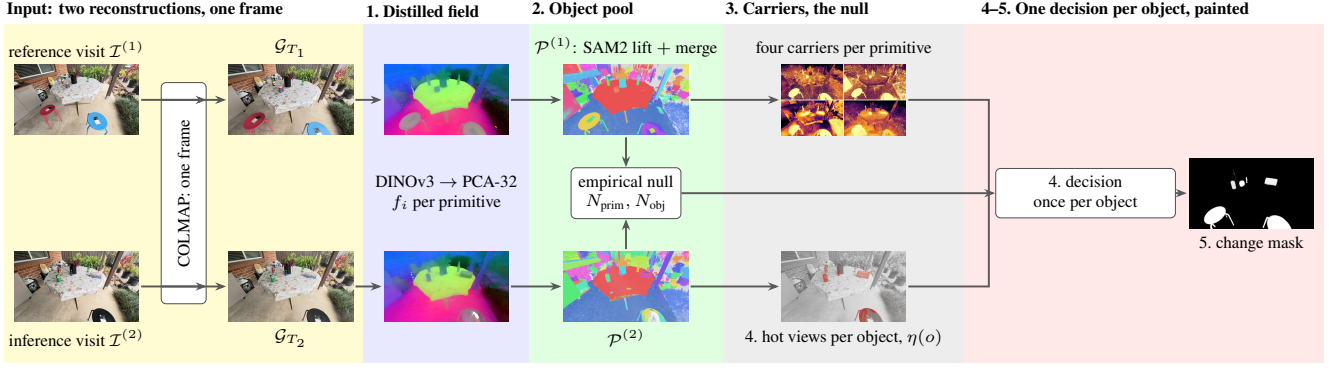}
\caption{\gspool{} on one PASLCD cell: a distilled DINOv3 field per visit (1), lifted SAM2 segments merged into objects (2), four carriers per primitive and the scene's null (3), one decision per object (4), and the change mask and changed primitives (5).}
\label{fig:pipeline}
\end{figure*}

\gspool{} takes two independently reconstructed Gaussian fields $\Gone$ and
$\Gtwo$ of the same space, sharing a camera frame, and returns the changed
objects in each together with their masks. These two reconstructions serve as inputs that pass sequentially
through five stages, allowing each change decision to be traced back through
the evidence behind it. Each stage is presented in this order below (Figure~\ref{fig:pipeline}). We refer
to $T_1$ as the reference visit and $T_2$ the inference visit, the visit for
which the change map is evaluated.

\textbf{Decisions in 3D.} \gspool{} labels objects formed from primitives. The photographs supply the segments, their tracking and the photographic evidence aggregated on the primitives, and they confirm or reject the changes proposed in 3D. The final mask renders the changed objects' selected primitives.

\textbf{Inputs.} Each visit is reconstructed independently by an unmodified
3DGS trainer at its published
defaults. We use vanilla 3DGS~\cite{3dgs} and FastPGSR, the FastGS~\cite{fastgs}
acceleration of PGSR~\cite{pgsr}. The two reconstructions share a camera frame, which \gspool{} takes as given.

\subsection{Stage 1: the distilled semantic field}

We attach a semantic descriptor to every primitive. DINOv3 (ViT-L/16) features
are extracted from the undistorted views, reduced by one PCA to $D=32$
dimensions, and distilled onto the primitives as in Feature
3DGS~\cite{feat3dgs}: a per-primitive feature vector $f_i \in \mathbb{R}^{32}$
is rendered by gsplat~\cite{gsplat} with the same alpha compositing as colour and optimised to match every view's upsampled patch features. The reconstruction itself is held fixed: splat positions, scales, rotations, opacities and colour are not optimised, only $f$ is. The PCA is fitted to both visits together, so the two visits' features lie in one feature space and can be compared directly.

\subsection{Stage 2: the object pool}
\label{sec:pool}

The pool turns primitives into objects: SAM2 segments each visit's
photographs and the segments are lifted onto the splat. Segments of the same
object are merged.

\textbf{Segments.} SAM2 runs on the
keyframes, a subset of each visit's photographs. On keyframe
$\ell$ it returns a label map, an image that gives each pixel the id of at most
one segment, and segment $m$ is the mask $M_{\ell,m}$ of the pixels with id $m$. The two visits
are segmented independently with approximately the same number of photographs per viewpoint: the visit with more photographs (usually $T_1$) is sub-sampled evenly along its capture to at most the number of the other visit's photographs, all of which are keyframes. This keeps the two object pools comparable.

\textbf{Lifting.} Each mask $M_{\ell,m}$ is lifted onto the primitives, that
is, assigned to the primitives that render it. We render keyframe $\ell$ with
the visit's own reconstruction, and the mask \emph{claims} primitive $i$ when the following
two tests hold. First, $i$ must lie on the rendered surface. With $e_i(\ell)$ the
difference between the depth of $i$'s centre and the rendered depth at its
pixel, $|e_i(\ell)|$ must be below three times the median of $|e_j(\ell)|$ over
the primitives that project into the keyframe. This tolerance keeps the primitives
that lie near the surface and belong to it. Second, at least
half of $i$'s weight in view $\ell$ must land inside the mask. With $a_{i,\ell,p}$ the weight the renderer assigns to $i$ at
pixel $p$, its alpha-composited contribution,
\begin{equation}
\phi_i(M_{\ell,m}) \;=\; \frac{\sum_{p \in M_{\ell,m}} a_{i,\ell,p}}{\sum_{p} a_{i,\ell,p}} \;\ge\; \tfrac{1}{2} .
\label{eq:liftshare}
\end{equation}
For example, a wall primitive whose edge brushes a chair's segment has little
weight inside the chair's mask and is not part of the
chair. Both sums are read off the rasteriser's backward pass:
FlashSplat~\cite{flashsplat} accumulates the same per-mask weights during
rasterisation.\ifalgorithmshown{} Algorithm~\ref{alg:gspool} lists both tests.\fi

\textbf{Merging.} Within a visit, SAM2's own video predictor carries each
keyframe's masks into the next keyframe, so a segment keeps its id along the
keyframe chain, and the segments that share an id form a \emph{track}. A chain breaks when an object leaves the view and returns, or when it is segmented from two sides that never share a keyframe, so one object can yield several tracks that claim the same primitives. Tracks are therefore grouped after lifting. Let $A$ and $B$ be the sets of primitives that two tracks claim. The two tracks form one object when these sets overlap,
\begin{equation}\label{eq:merge}
\mathrm{IoU}(A,B) \;=\; \frac{|A \cap B|}{|A \cup B|} \;\ge\; \tfrac{1}{2} .
\end{equation}
Grouping is greedy: among all pairs of tracks that claim a common primitive, it merges the pair with the highest IoU first, and it compares a merged group with the others through the union of its primitives, so a single overlapping pair cannot link two regions that do not overlap.

After merging, a primitive can still be claimed by more than one object: a
cup's rim may fall inside the cup's masks in some views and the table's in
others. A primitive belongs to the object whose masks hold most of its rendering weight, among the objects that claimed it in at least two keyframes. Two is the fewest claims that can agree, so a primitive that no object claimed at least twice stays unassigned.

The result is the object pool of visit $v$,
$\mathcal{P}^{(v)} = \{o_1,\dots,o_n\}$, a disjoint collection of primitive sets. A primitive is \emph{covered}
when it lies inside at least two camera frusta of the other visit $\bar v$.
Write $o_{\bar v}$ for the covered primitives of $o$ and
$|o|_{\bar v}$ for their number.
Two kinds of object are excluded. The smallest have too few primitives to aggregate evidence over. The largest is usually the space itself, because lifting fuses the floor, walls and ceiling into one surface. With $W$ the number of covered primitives in the objects at or above the size threshold, an object that holds more than half of $W$ is treated as background and removed. An object is
\emph{poolable} when
\begin{equation}
Q_{0.25}\bigl\{\,|o'|_{\bar v} : o' \in \mathcal{P}^{(v)}\bigr\} \;\le\; |o|_{\bar v} \;\le\; \tfrac{1}{2}\,W .
\label{eq:poolable}
\end{equation}
Here $o'$ runs over every object of the same pool, so the size threshold is the lower quartile of the pool's own object sizes. A fixed primitive count would not transfer, because the same object is a few dozen primitives in one reconstruction and several hundred in another. Appendix~\ref{sec:constants} sweeps the quantile and finds a broad plateau around the chosen value. An object that is not poolable is discarded. It receives no change decision and is not used to calibrate the decision threshold.

\subsection{Stage 3: the four carriers}
\label{sec:evidence}

Every primitive now receives four change scores, which we call \emph{carriers}. Three come from kernels that compare the two primitive sets (geometry, colour and semantics), and the fourth from the other visit's photographs. All four are on the same scale of $[0,1]$, and none depends on the object pool. A score of $1$ means that this carrier detected change in that primitive.

\textbf{Drift} (\gsdiff{}~\cite{gsdiff}). Two reconstructions of the same unchanged surface never coincide exactly, so every kernel tolerates some drift. As in \gsdiff{}, the drift is estimated from the scene itself as the upper-quartile distance from
a primitive to its nearest primitive in the other visit, and split in two: $u_n$
across the primitive's surface, along its normal, and $u_t$ within the surface.
Both lengths, together with the positional uncertainty of the primitive given the cameras, widen its covariance $\Sigma_i$ to the $\tilde\Sigma_i$ used by the kernels (Appendix~\ref{sec:gsdiffkernels}). All distances below are in units of $u_n$, the drift length.

\textbf{Search ball.} Each kernel compares a primitive $i$ of visit $v$,
centred at $\mu_i$, with nearby primitives $j$ of the other visit through the
offset $\Delta_{ij} = \mu_i - \mu_j$, and takes a \emph{maximum} over them, so $i$ counts as unchanged if any one of them matches it. A maximum grows with the number of
candidates: a primitive on a densely reconstructed bookshelf has many, one of
which may resemble it by chance, while the same primitive on a sparse wall has
few. Each primitive is therefore compared with at most $k$ primitives of the
other visit, its \emph{search ball} $\mathcal{B}(i)$: the $k$ nearest within three standard deviations of its widened covariance (\gsdiff's Eq.~6). The budget $k$ is the median number of the other visit's primitives within $3u_t$ of a primitive, so a dense region gets no more chances to match than a sparse one\ifalgorithmshown{} (Algorithm~\ref{alg:gspool})\fi.

\textbf{Geometry and colour} (\gsdiff{}~\cite{gsdiff}). The first two carriers
come from \gsdiff's kernels over this search ball (Appendix~\ref{sec:gsdiffkernels}).
The geometric mismatch $\delta^{\text{geo}}_i$ is low when some neighbour sits where $i$ does, with a similar shape. It is weighted by $\omega_i$, the confidence in $i$'s position given the cameras, so the geometric carrier is
$\omega_i\,\delta^{\text{geo}}_i$. The colour mismatch $\delta^{\text{col}}_i$
is low when some neighbour has $i$'s colour, up to the scene's colour scale $\sigma_{\text{col}}$. A large primitive gets a wider scale, multiplied by $s_i \ge 1$, its squared size relative to the visit's median.

\textbf{Semantics.} We extend the appearance-kernel construction to distilled
features, with a separate scale for each dimension,
\begin{subequations}\label{eq:dsem}
\begin{align}
d^2_{ij} &= \frac{1}{D}\sum_{d=1}^{D}\left(\frac{f_{i,d}-f_{j,d}}{\sigma_{f,d}}\right)^{\!2},
\label{eq:semdist}\\[2pt]
\delta^{\text{sem}}_i &= 1 - \max_{j \in \mathcal{B}(i)}\exp\!\left(-\frac{d^2_{ij}}{2 s_i}\right),
\label{eq:semres}
\end{align}
\end{subequations}
where $f_{i,d}$ is coordinate $d$ of primitive $i$'s distilled feature
$f_i \in \mathbb{R}^{D}$ (Stage~1) and $\sigma_{f,d}$ the scale of
that coordinate, measured from the same nearest-primitive pairs as $\sigma_{\text{col}}$ (Appendix~\ref{sec:gsdiffkernels}). The scale is per coordinate because the leading PCA components vary widely across the whole scene, and under a single scale they would dominate the distance and hide a small distinctive object.

\begin{figure}[t]
\centering
\includegraphics[width=\columnwidth]{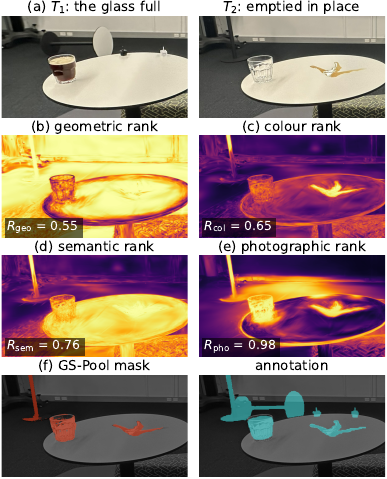}
\caption{A glass emptied in place on Lounge (a). Geometry and colour rank it near the scene's middle (b, c), semantics higher (d) and the photographs near the top (e). (f) \gspool{}'s mask and the annotation.}
\label{fig:photo}
\end{figure}

\textbf{Photographs.} The fourth carrier is read from the
photographs, because a photograph holds the appearance at capture time that
neither reconstruction keeps exactly. Each visit's model is rendered at the
camera of every photograph $\ell$ of the \emph{other} visit and compared with
it pixel by pixel,
\begin{subequations}\label{eq:cue}
\begin{align}
\mathcal{L}_1(p) &= \tfrac{1}{3}\bigl\lVert I_\ell(p) - \hat I_\ell(p)\bigr\rVert_1 ,
\label{eq:cuel1}\\[2pt]
\rho_\ell(p) &= 0.8\cdot\mathcal{L}_1(p) + 0.2\cdot\mathrm{DSSIM}\big(I_\ell, \hat I_\ell\big)(p),
\label{eq:cuerho}
\end{align}
\end{subequations}
where $p$ is a pixel, $I_\ell$ the photograph and $\hat I_\ell$ the model's render
from its camera, $\mathcal{L}_1$ the colour error averaged over the
three channels, $\tfrac13\bigl(|\Delta R| + |\Delta G| + |\Delta B|\bigr)$, and
${\mathrm{DSSIM} = \tfrac{1}{2}(1-\mathrm{SSIM})}$~\cite{ssim} the structural
dissimilarity on luma. Eq.~\ref{eq:cuerho} is the 3DGS training loss, evaluated per pixel. Training backpropagates this loss through the rasteriser to optimise the primitives. We backpropagate it in the same way, against the other visit's photographs, to score each primitive:
\begin{equation}
\delta^{\text{pho}}_i \;=\;
\frac{\sum_{\ell}\sum_{p} a_{i,\ell,p}\,\rho_\ell(p)}
     {\sum_{\ell}\sum_{p} a_{i,\ell,p}} ,
\label{eq:dpho}
\end{equation}
with $a_{i,\ell,p}$ the alpha-composited contribution of Eq.~\ref{eq:liftshare},
here at the other visit's cameras. One backward pass of the rasteriser per view computes both weighted sums, so the attribution respects occlusion. A primitive that contributes to no pixel of the other visit's photographs scores $\delta^{\text{pho}}_i = 0$. Figure~\ref{fig:photo} shows why the
carrier is needed. A glass emptied in place keeps its shape and its position,
so geometry and colour, which compare the two reconstructions with each other, cannot separate it from drift, while the photographic carrier still ranks it near the top. Because this loss is backpropagated to the primitives, it is aggregated per object and weighed against the other carriers, which a 2D change map cannot do. \mvcd{}, which compares each view in 2D before fusing, scores 36\% below \gspool{} on PASLCD (Table~\ref{tab:main}). On its own, the photographic carrier reaches 62\% of the full method's score on FastPGSR (Table~\ref{tab:rung}), and removing it costs 12.9\% on FastPGSR and 20.5\% on vanilla 3DGS, the largest loss of any carrier (Appendix~\ref{sec:ladder}).

\textbf{Empirical null.} The search ball makes carriers comparable between regions of a scene: unchanged
reconstructions disagree by an amount that varies with the scene, the trainer
and the carrier. Each visit therefore supplies its own set of primitives and objects present in both visits, which are assumed unchanged. We call this set the \emph{empirical null}, and its members null primitives and null objects. Let $j^\star$ be the primitive of the other visit
nearest to primitive $i$. Among the covered
primitives and poolable objects of visit $v$, those present in both visits are
\begin{subequations}\label{eq:null}
\medmuskip=2mu \thickmuskip=3mu
\begin{align}
N_{\text{prim}} &= \{i : \lVert\Delta_{ij^\star}\rVert < u_n\} ,
\label{eq:nullprim}\\[2pt]
N_{\text{obj}} &= \{o : |o_{\bar v} \cap N_{\text{prim}}| \ge \tfrac12 |o|_{\bar v} \vee \operatorname{med}_{o_{\bar v}} \lVert\Delta_{ij^\star}\rVert < 3u_n\} .
\label{eq:nullobj}
\end{align}
\end{subequations}
The test uses geometry alone, so an object changed in place can enter the null. Stage~4 ranks primitives against
$N_{\text{prim}}$ and objects against the other visit's $\bar N_{\text{obj}}$.

\subsection{Stage 4: the decision}
\label{sec:decision}

\begin{table*}[b]
\centering
\small
\setlength{\tabcolsep}{5pt}
\begin{tabular}{lllcc cccccc}
\toprule
& & & \multicolumn{2}{c}{mean} & \multicolumn{3}{c}{mIoU per reconstruction} & \multicolumn{3}{c}{F1 per reconstruction} \\
\cmidrule(lr){4-5}\cmidrule(lr){6-8}\cmidrule(lr){9-11}
Method & Frame & Annotations & mIoU & F1 & seed 22 & 23 & 24 & seed 22 & 23 & 24 \\
\midrule
\multicolumn{11}{@{}l}{\emph{PASLCD}} \\
\gspool{} (FastPGSR) & joint & PASLCD & \textbf{0.7509} & \textbf{0.8461} & 0.7404 & 0.7520 & 0.7602 & 0.8379 & 0.8466 & 0.8537 \\
\gspool{} (vanilla 3DGS) & joint & PASLCD & 0.7198 & 0.8209 & 0.7223 & 0.7111 & 0.7259 & 0.8236 & 0.8144 & 0.8248 \\
\gspool{} (vanilla 3DGS) & anchored & PASLCD & 0.7281 & 0.8266 & 0.7368 & 0.7139 & 0.7335 & 0.8341 & 0.8159 & 0.8300 \\
\gspool{} (FastPGSR) & anchored & PASLCD & 0.7260 & 0.8277 & 0.7135 & 0.7347 & 0.7300 & 0.8171 & 0.8351 & 0.8309 \\
\gsdiff{}~\cite{gsdiff} (FastPGSR) & joint & PASLCD & 0.644 & 0.758 & --- & --- & --- & --- & --- & --- \\
\oscd{}~\cite{oscd} & --- & PASLCD & 0.552 & 0.694 & --- & --- & --- & --- & --- & --- \\
PlenoCI~\cite{plenoci} & --- & PASLCD & 0.538 & 0.688 & --- & --- & --- & --- & --- & --- \\
\mvcd{}~\cite{mv3cd} & --- & PASLCD & 0.478 & 0.628 & --- & --- & --- & --- & --- & --- \\
\midrule
\multicolumn{11}{@{}l}{\emph{CL-Splats}} \\
PlenoCI~\cite{plenoci} & --- & PlenoCI & \textbf{0.951} & \textbf{0.975} & --- & --- & --- & --- & --- & --- \\
\gspool{} (vanilla 3DGS) & joint & \gspool{} & \textbf{0.8649} & \textbf{0.9237} & 0.8890 & 0.8500 & 0.8556 & 0.9395 & 0.9135 & 0.9180 \\
\gspool{} (FastPGSR) & joint & \gspool{} & 0.8550 & 0.9114 & 0.8755 & 0.8463 & 0.8431 & 0.9313 & 0.9039 & 0.8991 \\
\oscd{}~\cite{oscd} & --- & PlenoCI & 0.756 & 0.845 & --- & --- & --- & --- & --- & --- \\
\gsdiff{}~\cite{gsdiff} & --- & PlenoCI & 0.724 & 0.829 & --- & --- & --- & --- & --- & --- \\
\mvcd{}~\cite{mv3cd} & joint & \gspool{} & 0.6413 & 0.7667 & --- & --- & --- & --- & --- & --- \\
\oscd{}~\cite{oscd}$^\dagger$ & joint & \gspool{} & 0.6224 & 0.7428 & --- & --- & --- & --- & --- & --- \\
\mvcd{}~\cite{mv3cd} & --- & PlenoCI & 0.634 & 0.752 & --- & --- & --- & --- & --- & --- \\
\bottomrule
\end{tabular}
\caption{Change detection on PASLCD and CL-Splats. \gspool{} is given as the seed mean and per seed. Baselines are as reported (\mvcd{} on PASLCD as reported in~\cite{oscd}), except the rows on our CL-Splats annotations, which we ran ($^\dagger$four of five scenes).}
\label{tab:main}
\end{table*}

Stage~4 decides once per object whether it changed, in four steps.

\textbf{Object ranks.} The carriers' scales differ by carrier and scene. Each carrier $c \in \mathcal{C} = \{\text{col}, \text{geo}, \text{pho}, \text{sem}\}$ of Stage 3 becomes a scene-relative statistic in two steps: every primitive value $x^c_i$ (one of $\omega_i\delta^{\text{geo}}_i$, $\delta^{\text{col}}_i$, $\delta^{\text{pho}}_i$, $\delta^{\text{sem}}_i$) is replaced by its mid-rank $r^c_i$ among the same
visit's null primitives $N_{\text{prim}}$, and the object takes the mean of those ranks:
\begin{equation}
R_c(o) = \frac{1}{|o|_{\bar v}}\sum_{i \in o_{\bar v}} r_i^c .
\label{eq:pooledA}
\end{equation}
$R_c(o)$ is the normalised Mann--Whitney statistic~\cite{mannwhitney}: the
empirical probability that a covered primitive of $o$ has a larger
value on carrier $c$ than a null primitive. Values above
$\tfrac12$ indicate larger carriers on average. We average over the object to combine evidence spread over many primitives. More primitives per object give diminishing gains as the carriers are correlated.

\textbf{Core.} Each object is ranked against the other visit's null objects, and the decision threshold is set so that about one of those objects exceeds it. For a query $o$ of visit $v$, each $R_c(o)$ becomes its upper-tail rank
$p_c(o)$ among the other visit's null objects $\bar N_{\text{obj}}$,
with $\mathds{1}[\cdot]$ equal to $1$ when its condition holds and $0$
otherwise, and the four ranks are combined by Fisher's statistic~\cite{fisher}:
\begin{subequations}\label{eq:pvalue}
\begin{align}
p_c(o) &= \frac{1 + \sum_{o' \in \bar N_{\text{obj}}} \mathds{1}\bigl[R_c(o) \le R_c(o')\bigr]}{1 + |\bar N_{\text{obj}}|} ,
\label{eq:pvaluedef}\\[2pt]
F(o) &= -2\textstyle\sum_c \log p_c(o) .
\label{eq:fisherF}
\end{align}
\end{subequations}
For four independent uniform ranks $F$ follows $\chi^2_8$. A scaled
$\kappa\,\chi^2_\nu$, as in Brown's correction~\cite{brown} and its empirical version~\cite{ebm}
but with the mean and variance of the scores of $\bar N_{\text{obj}}$
(Appendix~\ref{sec:circularity}), accounts for the dependence between the carriers and the discreteness of the ranks. The core flags $o$ as changed when $F(o)$ exceeds
\begin{equation}
\theta \;=\; \kappa\,Q_{\chi^2_\nu}\!\left(1 - \frac{1}{|\bar N_{\text{obj}}|}\right),
\label{eq:corebar}
\end{equation}
with $Q_{\chi^2_\nu}$ the quantile function, so that about one of the $|\bar N_{\text{obj}}|$ null objects exceeds $\theta$. Replacing the carriers' average (row 3 of Table~\ref{tab:rung}) with this core (row 4) raises mIoU on both trainers.

\begin{figure*}[t]
\centering
\includegraphics[width=\textwidth]{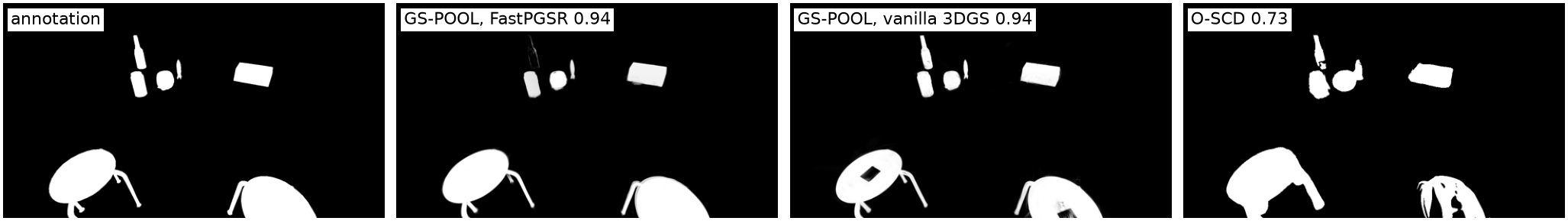}
\caption{Porch at the test view with the most annotated change: the annotation, \gspool{} on FastPGSR and on vanilla 3DGS, and \oscd{}, each tagged with its view IoU.}
\label{fig:qual}
\end{figure*}

\textbf{Photographic observation.}\label{sec:photo}
The core uses only an object's primitives, which inherit the instability of the reconstruction. It can therefore flag an unchanged object whose primitives drifted, and miss a change that covers few primitives. The photographs do not depend on the reconstruction, so we also test each object against them directly.

Take an object $o$ and a photograph $\ell$ of the other visit $\bar v$, and
render the reconstruction $o$ belongs to at $\ell$'s camera. The \emph{footprint} of $o$ in $\ell$ is the set of
pixels whose weight comes at least half from its covered primitives,
$\sum_{i \in o_{\bar v}} a_{i,\ell,p} \ge \frac12$, so pixels where other primitives occlude $o$ do not count. The photograph \emph{observes} $o$ when the
footprint is not empty, and is \emph{hot} for $o$ when the mean photographic
loss over the footprint, $\bar\rho_\ell(o)$ (Eq.~\ref{eq:cue}), exceeds
$Q_{0.95}\{\rho_\ell\}$, the 95th percentile of $\rho_\ell$ over the whole photograph
(Appendix~\ref{sec:constants} sweeps this quantile). With $\bar V(o)$ the photographs of
$\bar v$ that observe $o$, both statistics count photographs in $\bar V(o)$ and divide the count by $|\bar V(o)|$, so each is a share of these photographs, and an object photographed more often does not score higher for that reason alone. $\eta(o)$ is the share of them that are hot for $o$, that is, in which $o$ is among the photograph's highest losses, and $\gamma(o)$ is the share in which $o$'s mean loss exceeds that of the photograph's median observed object:
\begin{align}
\eta(o) &= \frac{1}{|\bar V(o)|}\sum_{\ell \in \bar V(o)} \mathds{1}\bigl[\bar\rho_\ell(o) > Q_{0.95}\{\rho_\ell\}\bigr] , \label{eq:hot}\\
\gamma(o) &= \frac{1}{|\bar V(o)|}\sum_{\ell \in \bar V(o)} \mathds{1}\bigl[\bar\rho_\ell(o) > \operatorname{med}_{o'}\bar\rho_\ell(o')\bigr] .
\label{eq:median}
\end{align}

\textbf{Decision rule.} The rule combines the core with the two photograph statistics. Each clause handles one failure case:
\begin{itemize}\setlength\itemsep{1pt}
\item \emph{Small object.} It has too few primitives for the core to flag it, so the photographs admit it.
\item \emph{Unchanged object whose primitives drift.} Its photographs match, so the photographs reject it.
\item \emph{Low-contrast object}, such as a thin sheet laid on a surface of its own colour. Its pixels barely change, but its semantic rank is high, so it is not rejected.
\end{itemize}
The core alone already exceeds the best prior result on both trainers (row 4, Table~\ref{tab:rung}), and the clauses raise both further (rows 5 and 6).

The \emph{quorum} holds, written $\mathrm{quorum}(o)$, when at most one of the four object ranks of $o$ falls below $\tfrac12$,
\begin{subequations}\label{eq:photo}
\begin{equation}
\bigl|\{\, c \in \mathcal{C} : R_c(o) < \tfrac12 \,\}\bigr| \le 1 .
\label{eq:quorum}
\end{equation}
We admit $o$, written $\mathrm{admit}(o)$, when the quorum holds, its photographic rank is at least $\tfrac12$, and either the core flags it or at least half of the photographs that observe it are hot:
\begin{equation}
R_{\text{pho}}(o) \ge \tfrac{1}{2} \;\wedge\; \mathrm{quorum}(o) \;\wedge\; \bigl[F(o) > \theta \;\vee\; \eta(o) \ge \tfrac{1}{2}\bigr] .
\label{eq:admit}
\end{equation}
We reject $o$, written $\mathrm{reject}(o)$, when at least one photograph observes it but it exceeds the median object in fewer than half of those photographs, unless its semantic rank is in the top 5\% (the \emph{semantic exemption}):
\begin{equation}
|\bar V(o)| > 0 \;\wedge\; \gamma(o) < \tfrac{1}{2} \;\wedge\; R_{\text{sem}}(o) < 0.95 .
\label{eq:reject}
\end{equation}
\end{subequations}
An object $o$ is changed when $\mathrm{admit}(o) \wedge \neg\,\mathrm{reject}(o)$.

\subsection{Stage 5: changed primitives and masks}

\begin{figure*}[t]
\centering
\includegraphics[width=\textwidth]{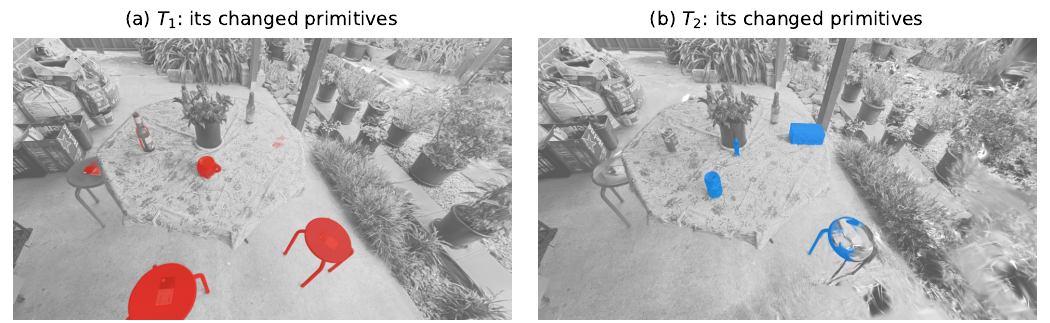}
\caption{The output in 3D on Porch from a viewpoint no camera took. Each visit's changed primitives are shown on its own reconstruction and the rest in grey: $T_1$'s two stools and mug (a), and $T_2$'s stool and the objects on its table (b).}
\label{fig:output3d}
\end{figure*}

A changed object's \emph{extent} is the set of its covered primitives that exceed the 95th percentile of $N_{\text{prim}}$ on at least one carrier,
\begin{equation}
E(o) \;=\; \bigl\{\, i \in o_{\bar v} \;:\; \exists c,\; x^c_i \;>\; Q_{0.95}\{x^c_j\}_{j \in N_{\text{prim}}} \,\bigr\} .
\label{eq:extent}
\end{equation}
The extents are rendered into each test view to form the change mask. Appendix~\ref{sec:constants} sweeps the quantile. Taking every covered primitive as the extent instead lowers mIoU on both trainers (Appendix~\ref{sec:ladder}), because an object's primitive set also includes primitives of the surfaces it touches, which did not change. Both visits are processed by the same rule, so each returns its own changed objects (Figure~\ref{fig:output3d}).

\section{Experiments}

\subsection{Setup}
\label{sec:protocol}

We evaluate on the 20 cells of PASLCD, each one instance of a scene, and on our annotations of CL-Splats, with the benchmark's own evaluator. A test view without a registered pose receives an empty mask. Every configuration runs over three independent reconstructions of each visit, giving $20 \times 3 = 60$ runs per table row and 240 in all. We call each run a \emph{member}, and we score members separately.
We report both trainers, because a detector's result depends on the
reconstructions it is given. The shared camera frame comes in two forms. The
\emph{joint} frame is the benchmark's own solve over both captures. The
\emph{anchored} frame registers the inference capture into the reference poses
without re-optimising them, a method more likely to be used in the industry, keeping the two captures truly independent.
(Appendix~\ref{sec:frames}).
Given the reconstructions, distilled fields and segmentation maps, one member
takes a median 2.1 minutes with FastPGSR and 5.7 with vanilla 3DGS on one RTX
4090 on the joint frame (Appendix~\ref{sec:compute}).

\subsection{Results}

All percentages in the text are relative changes in mIoU. On the joint frame \gspool{} exceeds the best prior result on
PASLCD by 17\% with FastPGSR and 12\% with vanilla 3DGS
(Table~\ref{tab:main}, Figure~\ref{fig:qual}).

\textbf{Stability.} Results vary little with the reconstruction seed. Across the three seeds, the standard deviation of a row's per-seed mean is at most 1.7\% of the row mean, and each of the six per-seed means on the joint frame exceeds the best prior result.
On the joint frame FastPGSR scores 4.3\% above vanilla 3DGS at every seed. Resampling the ten scenes gives a 95\% interval of 0.2\% to 9.1\% for this gain, so it is above zero but its size is uncertain (Appendix~\ref{sec:percell}). On the anchored frame the two trainers differ by only 0.3\%.

\textbf{CL-Splats.} On the real scenes of CL-Splats~\cite{ackermann2025clsplats},
annotated with PASLCD's convention, \gspool{} with the same configuration scores 33\% (FastPGSR) and 35\% (vanilla 3DGS) above \mvcd{} on the same masks. \oscd{} fails on one desk image, and on the other four scenes \gspool{} scores 35\% and 37\% above \oscd{} (Table~\ref{tab:main}, Appendix~\ref{sec:clsplats}). PlenoCI~\cite{plenoci} reports about 10\% more on this easier benchmark, on its own annotations and with thresholds selected per scene. On PASLCD, the harder one, \gspool{} scores 40\% above PlenoCI.

\subsection{Ablation}
\label{sec:ablation}

Table~\ref{tab:rung} adds one component of the decision at a time. Rows 1--3 admit an object whose score exceeds the $1-\frac{1}{|\bar N_{\text{obj}}|}$ quantile of the scores of the other visit's null objects. Every row raises mIoU on both trainers. Adding semantics to the photographs is the largest single step. Averaging all four carriers adds little, while combining them through the core adds a large step. The photographic-rank and quorum conditions (row 5) and the per-photograph statistics $\eta$ and $\gamma$, the \emph{view statistics} (row 6), complete the rule. Removing the rejection clause (Eq.~\ref{eq:reject}) gives an equal or lower score on each of the 120 joint members. In the controlled comparisons (Appendix~\ref{sec:ladder}), removing any carrier lowers the score. The photographs contribute most, adding 14.8\% on FastPGSR and 25.8\% on vanilla 3DGS with the same pools and rule. Deciding per object scores above deciding
per primitive. The size threshold and the extent quantile lie on broad plateaus, the view quantile's chosen $0.95$ is the best
value on both trainers, and SAM2's stability threshold costs at most 1.9\%
over its sweep (Appendices~\ref{sec:constants} and~\ref{sec:models}).

\begin{table}[ht]
\centering\small\setlength{\tabcolsep}{2.2pt}
\begin{tabular}{@{}clcccc@{}}
\toprule
& & \multicolumn{2}{c}{FastPGSR} & \multicolumn{2}{c}{vanilla 3DGS} \\
\cmidrule(lr){3-4}\cmidrule(lr){5-6}
row & decision & mIoU & F1 & mIoU & F1 \\
\midrule
1 & photographs alone & 0.4681 & 0.5896 & 0.4353 & 0.5627 \\
2 & \quad + semantics & 0.5899 & 0.7087 & 0.6060 & 0.7291 \\
3 & all four carriers, averaged & 0.6090 & 0.7252 & 0.6107 & 0.7357 \\
4 & core, $F(o) > \theta$ & 0.7034 & 0.8080 & 0.6617 & 0.7762 \\
5 & + photo rank, quorum & 0.7113 & 0.8147 & 0.6668 & 0.7796 \\
6 & + view statistics (full rule) & \textbf{0.7509} & \textbf{0.8461} & \textbf{0.7198} & \textbf{0.8209} \\
\bottomrule
\end{tabular}

\caption{The decision rule built up one component at a time on the same pools and carriers. Mean over the 60 joint-frame members per trainer. Row 6 is the full rule.}
\label{tab:rung}
\end{table}

\section{Limitations}
\label{sec:limits}

\gspool{} takes all its evidence from the reconstructions and photographs, so a poor capture degrades it.
\gspool{}'s objects come from SAM2, so a change that SAM2 does not segment as a separate mask cannot be recovered by later stages. A change of lighting alone still produces false alarms, although on two captures of an unchanged scene under different lighting \gspool{} flags fewer pixels than \mvcd{} and \oscd{} run from their released code (Appendix~\ref{sec:unchanged}). The distilled field, SAM2 lifting and renders
at the other visit's cameras also make \gspool{} costlier than comparing the two primitive sets alone.

\section{Conclusion}

\gspool{} detects change per object between two independently trained 3DGS reconstructions. It computes four carriers per primitive, geometric, colour, semantic and photographic (the last backpropagated through the rasteriser), and ranks each object against the other visit's null objects. With one configuration across two trainers, two camera frames and three reconstructions each, \gspool{} reaches 0.7509 mIoU / 0.8461 F1 on PASLCD, 17\% above the best prior result, and 0.855 / 0.865 mIoU (FastPGSR / vanilla 3DGS) on CL-Splats. Between two captures of an unchanged scene under different lighting, it flags fewer pixels than \mvcd{} and \oscd{}. Its output is
each visit's object pool, every object with its decision and its evidence, and the changed primitives as a 3DGS layer over either reconstruction. \gspool{} therefore takes the problem of detection from individual primitives to whole objects, which opens the way to classifying the kind of change each object underwent, which we will explore further in the future.

\clearpage
\maketitlesupplementary
\appendix
\raggedbottom

\section{\gsdiff{} kernels}
\label{sec:gsdiffkernels}

The geometric and colour carriers of Stage~3 are \gsdiff's kernels~\cite{gsdiff},
run on the input reconstructions over the search ball of Stage~3\ifalgorithmshown{} (Algorithm~\ref{alg:gspool})\fi. This section gives them in full.

\textbf{Drift and widened covariances} (\gsdiff{}~\cite{gsdiff}, Eqs.~1--5). Two
reconstructions of the same unchanged surface never coincide exactly, so every
kernel allows for a drift measured on the scene: the 75th percentile of the
offset from a primitive to its nearest primitive in the other visit, taken in
each direction and combined as their root mean square. The offset splits into a
part along the primitive's normal, $u_n$, and a part in its surface, $u_t$,
because the photographs constrain depth more tightly than position along a surface. The two lengths, together with each position's uncertainty given the cameras, widen every covariance $\Sigma_i$ to the $\tilde\Sigma_i$ used by the kernels (\gsdiff's
Eqs.~3--5). Colour and semantics take their scales $\sigma_{\text{col}}$ and
$\sigma_{f,d}$ from the same pairs, the difference to the nearest primitive.
For geometry and colour, the drift is measured on the primitives that at least one camera of the other visit sees, as in \gsdiff's Algorithm~1. For semantics and the null (Eq.~\ref{eq:null}), it is measured on the covered primitives.

\textbf{Size-dependent scale} (\gsdiff{}~\cite{gsdiff}, Eq.~10). A large
primitive averages more surface, so its colour and features are compared more
loosely,
\begin{equation*}
s_i \;=\; \max\!\left(\frac{h_i^2}{\tilde h^2},\;1\right), \qquad h_i^2 = \mathrm{tr}\,\tilde\Sigma_i ,
\tag{\gsdiff{}~10}\label{eq:si}
\end{equation*}
with $h_i^2$ the primitive's squared size and $\tilde h^2$ its median over the
visit.

\textbf{Geometry} (\gsdiff{}~\cite{gsdiff}). The geometric mismatch is low when some neighbour sits where $i$ does,
\begin{equation*}
\delta^{\text{geo}}_i = 1 - \max_{j \in \mathcal{B}(i)} \exp\!\left(-\tfrac{1}{2}\,
  \Delta_{ij}^{\top}\big(\tilde\Sigma_i + \tilde\Sigma_j\big)^{-1} \Delta_{ij}\right),
\tag{\gsdiff{}~8}\label{eq:dgeo}
\end{equation*}
a Mahalanobis kernel~\cite{mahalanobis} whose covariance is the sum of the two widened covariances. Only this mismatch is weighted, by the confidence in the primitive's position,
\begin{equation*}
\omega_i \;=\; \sigma\!\left(\log \mathrm{tr}(H_i) - \log Q_{0.25}\{\mathrm{tr}(H_j)\}_{j \in \mathcal{S}}\right),
\tag{\gsdiff{}~13}\label{eq:omega}
\end{equation*}
with $\sigma$ the logistic function, $H_i$ the Fisher
information~\cite{jiang2024fisherrf} of $i$'s position over the cameras that saw
it and $Q_{0.25}$ its lower quartile over $\mathcal{S}$, the primitives of $v$
that at least one camera of $\bar v$ sees. The geometric carrier is $\omega_i\,\delta^{\text{geo}}_i$.

\textbf{Colour} (\gsdiff{}~\cite{gsdiff}). With $c_i$ a primitive's diffuse colour
(its zeroth spherical-harmonic band),
\begin{equation*}
\delta^{\text{col}}_i = 1 - \max_{j \in \mathcal{B}(i)}
  \exp\!\left(-\frac{\lVert c_i - c_j\rVert^2}{2\,\sigma_{\text{col}}^2 s_i}\right),
\tag{\gsdiff{}~12}\label{eq:dcol}
\end{equation*}
a Gaussian on the colour difference at the scale $\sigma_{\text{col}}$, widened for a large
primitive by $s_i$.

\section{Pool record}
\label{sec:poolrecord}

{\raggedright
Each member writes a pool record, \texttt{pool.json}. Per visit it lists the object counts and the core threshold $\theta$ (\texttt{core\_threshold}), and per poolable object every quantity its decision uses.
\texttt{covered\_primitives} is $|o|_{\bar v}$, \texttt{carrier\_ranks} the four
ranks $R_c(o)$ (Eq.~\ref{eq:pooledA}), \texttt{core\_statistic} $F(o)$
(Eq.~\ref{eq:fisherF}), \texttt{photographs\_observing} $|\bar V(o)|$,
\texttt{hot\_view\_share} $\eta(o)$ and \texttt{above\_median\_share} $\gamma(o)$
(Eqs.~\ref{eq:hot} and~\ref{eq:median}), \texttt{carriers\_agree} the quorum
$\mathrm{quorum}(o)$ (Eq.~\ref{eq:quorum}), \texttt{rejected\_by\_photographs}
$\mathrm{reject}(o)$ (Eq.~\ref{eq:reject}), and \texttt{extent\_primitives} the size of the extent $E(o)$ of a changed object (Eq.~\ref{eq:extent}). Object 31 passes every clause. Object 3 is far below the core threshold, and two of its ranks fall below $\tfrac12$.
\par}

{\fontsize{6}{7}\selectfont
\begin{verbatim}
{
  "member": "Garden_i1, FastPGSR, joint frame, seed 22",
  "gspool_version": "1.2",
  "T1": {"objects": 170, "poolable_objects": 126,
   "changed_objects": 9,
   "core_threshold": 22.1468,
   "pool": [
    {"object_id": 31, "covered_primitives": 534,
     "extent_primitives": 534,
     "carrier_ranks": {"colour": 0.9889, "geometry": 0.9993,
                       "photographs": 0.9995, "semantics": 0.9981},
     "core_statistic": 36.1743, "core_admits": true,
     "photographs_observing": 25, "hot_view_share": 1.0,
     "above_median_share": 1.0, "carriers_agree": true,
     "semantic_exemption": true, "admitted": true,
     "rejected_by_photographs": false, "changed": true},
    ...
    {"object_id": 3, "covered_primitives": 8100,
     "extent_primitives": 3389,
     "carrier_ranks": {"colour": 0.4059, "geometry": 0.7603,
                       "photographs": 0.7226, "semantics": 0.089},
     "core_statistic": 3.5916, "core_admits": false,
     "photographs_observing": 25, "hot_view_share": 0.0,
     "above_median_share": 0.96, "carriers_agree": false,
     "semantic_exemption": false, "admitted": false,
     "rejected_by_photographs": false, "changed": false},
    ...
   ]},
  "T2": {"objects": 168, "poolable_objects": 125,
   "changed_objects": 1,
   "core_threshold": 23.8048,
   "pool": [
    {"object_id": 11, "covered_primitives": 413,
     "extent_primitives": 413,
     "carrier_ranks": {"colour": 0.9988, "geometry": 0.9999,
                       "photographs": 0.9997, "semantics": 0.981},
     "core_statistic": 36.431, "core_admits": true,
     "photographs_observing": 60, "hot_view_share": 1.0,
     "above_median_share": 1.0, "carriers_agree": true,
     "semantic_exemption": true, "admitted": true,
     "rejected_by_photographs": false, "changed": true},
    ...
   ]}
}
\end{verbatim}
}

\clearpage

\begin{strip}\noindent\begin{minipage}{\textwidth}\centering\captionsetup{hypcap=false}
\includegraphics[width=0.92\textwidth]{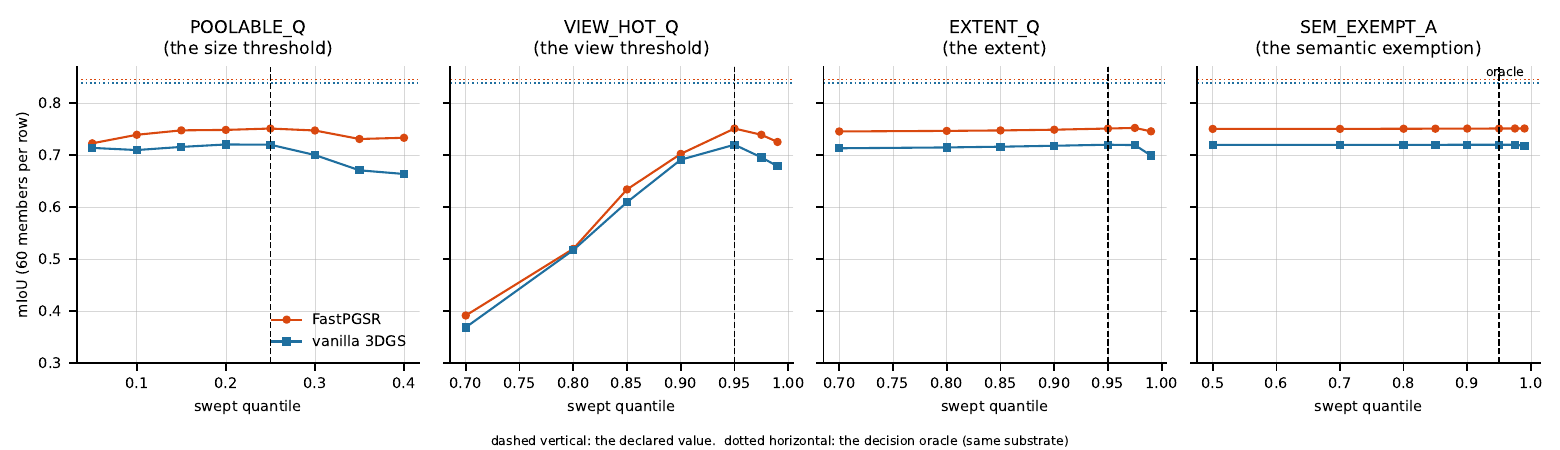}
\captionof{figure}{The four quantile constants, each swept with the other three fixed. Mean mIoU over the 60 members of each joint-frame row. Dashed: the chosen value. Dotted: the decision oracle (Appendix~\ref{sec:oraclebounds}).}
\label{fig:quantile}
\end{minipage}\end{strip}

\section{Constants}
\label{sec:constants}

We use one configuration across scenes.

\textbf{Quantiles} $\mathrm{POOLABLE\_Q}=0.25$ (size threshold,
Eq.~\ref{eq:poolable}), $\mathrm{VIEW\_HOT\_Q}=0.95$ (view threshold,
Sec.~\ref{sec:photo}), $\mathrm{EXTENT\_Q}=0.95$ (extent,
Eq.~\ref{eq:extent}) and $\mathrm{SEM\_EXEMPT\_A}=0.95$ (semantic exemption,
Sec.~\ref{sec:photo}) are quantiles of the scene's own distributions, so each adapts to the scene instead of fixing an absolute value.
Figure~\ref{fig:quantile} sweeps each with the other three fixed, on all 120
joint members. The size threshold, the extent quantile and the semantic exemption lie on broad plateaus around the chosen values. The view threshold is the sharpest, and its chosen $0.95$ is the best value on both trainers. No swept value reaches the decision oracle's score (Appendix~\ref{sec:oraclebounds}).

\section{Model settings}
\label{sec:models}

\begin{figure}[H]\centering
\includegraphics[width=\columnwidth]{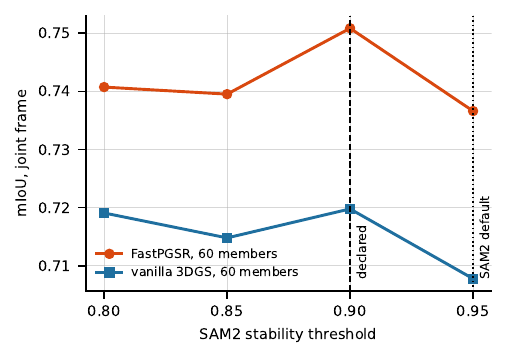}
\caption{SAM2's stability threshold on the joint frame, mean over 60 members per trainer. Dashed: the chosen $0.90$. Dotted: the library's default $0.95$.}
\label{fig:sam2}
\end{figure}

\textbf{SAM2.} We use \texttt{facebook/sam2-hiera-large}, frozen, through its
automatic mask generator with 32 points per side, a stability threshold of
$0.90$ and the library's other defaults, on the keyframes of Section~\ref{sec:pool}. Mask ids are linked between keyframes by the library's video predictor at its working resolution of 1024 pixels. No prompt uses any change information. The stability threshold decides which of SAM2's masks are kept, and so which segments the pool starts from. For each value of it we regenerate the label maps and rerun every member (Figure~\ref{fig:sam2}). The chosen $0.90$ is the best value on both trainers, and the generator's default $0.95$ costs 1.9\% on FastPGSR and 1.7\% on vanilla 3DGS.

\textbf{DINOv3 and distillation.} ViT-L/16, frozen, reads the whole undistorted
view resized to a long side of $1184$ pixels (74 patches). One exact PCA over the patch features of both visits reduces them to $D=32$. Only the
per-primitive feature vector is trained, for 5{,}000 steps against these
targets with the reconstruction held fixed. The reconstruction's seed initialises the feature vectors and selects the training view at each step.

\textbf{Trainers.} Both run at their published defaults with spherical
harmonics of degree 3: vanilla 3DGS, the reference implementation, for
30{,}000 iterations with densification from iteration 500 to 15{,}000, and
FastPGSR likewise, its planar and multi-view terms on from iteration 7{,}000. A row's three members differ only in the seed (22, 23, 24).

\section{Camera frames}
\label{sec:frames}

The detector does no registration. The two visits' cameras are given in one coordinate frame. Write $\mathcal{I}^{(1)}$ and $\mathcal{I}^{(2)}$ for the photographs of
the reference and the inference visit, $P_\ell$ for the pose of photograph
$\ell$, $X_i$ for a scene point, $x_{\ell i}$ for its observation in $\ell$, $\pi$
for the projection, $\psi$ for a robust loss and
$\mathcal{I} = \mathcal{I}^{(1)} \cup \mathcal{I}^{(2)}$.

\textbf{Joint.} PASLCD's own COLMAP solve~\cite{colmap}, the frame the
benchmark's baselines report, is one bundle adjustment over both visits that
moves every camera and every point:
\begin{equation}
\min_{\{P_\ell\}_{\ell \in \mathcal{I}},\; \{X_i\}} \;
\sum_{\ell \in \mathcal{I}} \sum_{i} \psi\bigl(\lVert \pi(P_\ell, X_i) - x_{\ell i} \rVert^2\bigr) .
\label{eq:jointba}
\end{equation}

\textbf{Anchored.} This is the frame available in deployment, where a new capture of a site is compared with an archived capture whose cameras are not solved again. The reference visit is solved
alone, Eq.~\ref{eq:jointba} over $\mathcal{I}^{(1)}$ only, giving its poses $P^{(1)}_\ell$. Its points are re-triangulated with those poses held, and the inference visit is registered into that model with every reference camera
fixed:
\begin{equation}
\min_{\{P_\ell\}_{\ell \in \mathcal{I}^{(2)}}} \;
\sum_{\ell \in \mathcal{I}^{(2)}} \sum_{i} \psi\bigl(\lVert \pi(P_\ell, X_i) - x_{\ell i} \rVert^2\bigr),
\label{eq:anchoredba}
\end{equation}
with $P_\ell = P^{(1)}_\ell$ held for every $\ell \in \mathcal{I}^{(1)}$.
The reference is never re-optimised, so every registration error belongs to the inference visit. The detector sees such an error only through larger nearest-neighbour offsets, which enlarge the drift lengths and the widened covariances and change the null. The
anchored frame changes FastPGSR's mean by $-3.3\%$ and vanilla 3DGS's by
$+1.2\%$, the latter inside the seed-to-seed spread (Table~\ref{tab:main},
Appendix~\ref{sec:percell}).

\section{Compute}
\label{sec:compute}

Every member was computed on NVIDIA RTX 4090 GPUs with 24\,GB from its stored reconstruction. Every member's outputs and run record are released~\cite{gspoolpaslcd, gspoolclsplats}.
Table~\ref{tab:timing} gives the median wall-clock per member.

\begin{table}[H]\centering\scriptsize\setlength{\tabcolsep}{3pt}
\begin{tabular}{@{}lrr@{}}
\toprule
row & prims/visit & time \\
\midrule
FastPGSR, joint & 0.19M & 2.1 min \\
FastPGSR, anchored & 0.17M & 1.7 min \\
vanilla 3DGS, joint & 1.14M & 5.7 min \\
vanilla 3DGS, anchored & 1.13M & 4.2 min \\
\bottomrule
\end{tabular}
\caption{Median wall-clock per member on one RTX 4090 (24\,GB), from the reconstructions, distilled fields, cached features and segmentation maps to the masks and change PLYs. Producing those inputs is excluded. Stage~3's neighbour searches run on the GPU.}
\label{tab:timing}
\end{table}

\section{Oracle scores}
\label{sec:oraclebounds}

The decision oracle uses the annotations to choose which objects to mark as changed. It starts from the full rule's decisions and makes one greedy pass that adds objects, then one that removes them, in object-id order over both visits. Each gain is evaluated on the union of the objects' individually thresholded rendered masks. The selected set is then rendered and scored as in Stage~5. This annotation-assisted search provides an achieved
score, not a proven optimum over object subsets.

Two pixel-level bounds take the masks of the same 60 FastPGSR joint members and either remove every false-positive pixel or add every missed pixel, at the benchmark's fixed evaluation threshold.

\begin{table}[H]
\centering\small
\begin{tabular}{lcc}
\toprule
& mIoU & F1 \\
\midrule
\gspool{}, re-scored by this pass          & 0.7509 & 0.8461 \\
\midrule
\emph{oracle:} greedy decisions, same pool      & \textbf{0.8448} & \textbf{0.9090} \\
\emph{oracle:} every false pixel removed        & 0.8172 & 0.8882 \\
\emph{oracle:} every missed pixel filled        & 0.9229 & 0.9568 \\
\bottomrule
\end{tabular}
\caption{The decision oracle and the two pixel bounds on the same 60 members (FastPGSR, joint frame). The first row re-scores \gspool{} with the same pass and reproduces its score exactly.}
\label{tab:oracle}
\end{table}

Filling every missed pixel gains 22.9\%. Removing every false pixel gains 8.8\%, less than the 12.5\% from greedy object selection. Missed changes and object decisions both leave room for improvement, but these bounds do not attribute each remaining error to the object pool or to the decision rule.

\section{Controlled comparisons}
\label{sec:ladder}
\begin{table*}[t]
\centering\small\setlength{\tabcolsep}{8pt}
\begin{tabular}{@{}llcccc@{}}
\toprule
& & \multicolumn{2}{c}{FastPGSR} & \multicolumn{2}{c}{vanilla 3DGS} \\
\cmidrule(lr){3-4}\cmidrule(lr){5-6}
& comparison & mIoU & F1 & mIoU & F1 \\
\midrule
& full rule & \textbf{0.7509} & \textbf{0.8461} & \textbf{0.7198} & \textbf{0.8209} \\
& core alone (row 4) & 0.7034 & 0.8080 & 0.6617 & 0.7762 \\
\midrule
null & full rule, own-visit null & 0.7464 & 0.8407 & 0.7269 & 0.8205 \\
 & full rule, both visits' null & 0.7357 & 0.8301 & 0.7195 & 0.8170 \\
 & core alone, own-visit null & 0.6530 & 0.7647 & 0.5986 & 0.7109 \\
 & core alone, both visits' null & 0.6298 & 0.7454 & 0.6130 & 0.7325 \\
\midrule
carriers & without geometry & 0.7098 & 0.8128 & 0.7174 & 0.8196 \\
 & without colour & 0.7273 & 0.8236 & 0.7019 & 0.8052 \\
 & without semantics & 0.7013 & 0.8044 & 0.6989 & 0.8044 \\
 & without photographs & 0.6541 & 0.7680 & 0.5721 & 0.6989 \\
\midrule
rejection & without the view rejection & 0.7493 & 0.8450 & 0.7195 & 0.8208 \\
 & without the semantic exemption & 0.7512 & 0.8463 & 0.7176 & 0.8194 \\
\midrule
unit & per primitive, four-carrier mean & 0.5070 & 0.6478 & 0.5557 & 0.6939 \\
 & every covered primitive as extent & 0.7443 & 0.8415 & 0.7120 & 0.8152 \\
\bottomrule
\end{tabular}

\caption{Twelve controlled comparisons around the full rule, each changing one thing, over the 120 joint members.}
\label{tab:comparisons}
\end{table*}

Table~\ref{tab:comparisons} changes one component of the full rule at a time on the same 120 joint members. It tests choices that the ablation of Table~\ref{tab:rung} keeps fixed.

\textbf{Null source.} A visit's objects are ranked against the other visit's null objects, so the threshold does not depend on the objects being judged. Using the visit's own null, or both visits' together, changes the full rule by $-2.0\%$ to $+1.0\%$, but lowers the core alone (row 4) by 7.2\% to 10.5\% on both trainers.

\textbf{Carriers.} Each carrier is dropped from both the decision and the extent. Every drop lowers the score on both trainers, so each carrier holds evidence that the other three miss. The photographs matter most.

\textbf{Unit.} In the per-primitive variant, each primitive's mean rank over the four carriers is thresholded at the $1-\frac{1}{|N_{\text{prim}}|}$ quantile of the primitive null, and an object is admitted if any of its primitives is selected. It scores below the same average pooled per object (row 3) on both trainers, so the decision is taken per object. Taking every covered primitive of a changed object as its extent, instead of Eq.~\ref{eq:extent}, lowers the score slightly on both trainers.

\textbf{Rejection.} The view rejection removes unchanged objects whose primitives drifted. Removing it lowers the score of 7 members on each trainer and raises none. The semantic exemption, which keeps a low-contrast change from being rejected, has a small effect of opposite sign on the two trainers.

\textbf{Row 5 conditions.} Under the core alone, the quorum changes the decision for only one object across the 120 members, so row 5's gain comes from the condition $R_{\text{pho}}(o) \ge \tfrac12$. In the full rule, the quorum excludes 131 objects, and the oracle would admit only 6 of them.

\section{Calibration across visits}
\label{sec:circularity}

An object's four ranks are dependent, because the four carriers are read on the same primitives, and they are discrete, because each is computed against a finite null. Fisher's statistic therefore does not follow $\chi^2_8$. Brown's correction~\cite{brown} and its empirical version~\cite{ebm} replace it with a scaled $\kappa\,\chi^2_\nu$ of the same mean and variance, taking the mean from theory and the variance from the correlations between the tests. We take both from the scores of the null objects, which also accounts for the discreteness of the ranks. With
$\mu_F$ and $\sigma^2_F$ the mean and sample variance of the scores of $\bar N_{\text{obj}}$,
each scored against that set including itself, matching
$\kappa \nu = \mu_F$ and $2\kappa^2 \nu = \sigma^2_F$ gives
\begin{equation}
\kappa = \frac{\sigma^2_F}{2\mu_F}, \qquad \nu = \frac{2\mu_F^2}{\sigma^2_F} .
\label{eq:brown}
\end{equation}
Four independent uniform ranks give $\kappa=1$ and $\nu=8$, Fisher's own $\chi^2_8$, and four identical ranks give $\kappa=4$ and $\nu=2$, so the fit moves from the first case to the second as the carriers agree more. Each visit's threshold uses only the other visit's null objects. The fit is approximate, and it does not guarantee a calibrated tail for a small null set or for a null set that contains an object changed in place.

\section{Unchanged scenes}
\label{sec:unchanged}

\textbf{Lighting pairs.} PASLCD's two instances of a scene share their changes
and their 25 test photographs under different lighting, so their two reference
captures are the same pre-change scene in two sessions. We reconstruct each at
seeds 22, 23 and 24 and transform instance 2's reconstruction into instance 1's frame by the similarity transform that aligns the shared test cameras (centre residual at most 0.28\% of the cameras' spread). The transformed model renders the original's views at a PSNR of 74.7\,dB or more. Every pixel the pair flags is a false positive.
Table~\ref{tab:lighting} reports the flagged share at instance 2's reference cameras, where \mvcd{} and \oscd{}, run from their released code with instance 1's photographs as reference, make their decisions on the same photographs. \oscd{}'s refinement fails on one scene. \gspool{} flags fewer pixels than both methods on average, and its median is far lower, because most of its false positives come from two scenes, one outdoors and one lit by a strong coloured light in its second capture.

\begin{table}[H]\centering\footnotesize\setlength{\tabcolsep}{4pt}
\caption{Pixels flagged between two captures of the same unchanged scene under different lighting (\%), at instance 2's reference cameras ($^\dagger$nine of ten scenes).}
\label{tab:lighting}
\begin{tabular}{@{}lccc@{}}
\toprule
Method & mean & median & scenes below \mvcd{} \\
\midrule
\mvcd{}~\cite{mv3cd} & 1.72 & 1.34 & --- \\
\oscd{}~\cite{oscd}$^\dagger$ & 1.47 & 1.09 & 3 of 9 \\
\gspool{} (FastPGSR) & 1.03 & 0.31 & 6 of 10 \\
\gspool{} (vanilla 3DGS) & 1.15 & 0.17 & 7 of 10 \\
\bottomrule
\end{tabular}
\end{table}

\textbf{Same-capture pairs.} This test has the design PlenoCI describes~\cite{plenoci}. Each
cell's reference capture is reconstructed at seeds 22, 23 and 24, and each of
the three pairs runs through the pipeline as two visits with the same photographs, cameras and label maps. Every pixel the change masks flag at the cell's 25 test views is a false positive. On the benchmark's real changes the
annotated masks cover 3.51\% of the pixels~\cite{plenoci}.

\begin{table}[H]\centering\small
\caption{Pixels flagged as changed between two reconstructions of an unchanged scene (\%). $^*$As reported by PlenoCI~\cite{plenoci} on its own reconstructions, not measured on our pairs.}
\label{tab:unchanged}
\begin{tabular}{@{}lccc@{}}
\toprule
Method & mean & median & runs at zero \\
\midrule
\gsdiff{}~\cite{gsdiff}$^*$ & 0.389 & --- & --- \\
PlenoCI~\cite{plenoci}$^*$ & 0.004 & --- & --- \\
\gspool{} (FastPGSR) & 0.153 & 0.017 & 14 of 60 \\
\gspool{} (vanilla 3DGS) & 0.511 & 0.017 & 15 of 60 \\
\bottomrule
\end{tabular}
\end{table}

The detector marks few objects on these pairs, 1.93 per visit on average with FastPGSR and 2.15 with vanilla 3DGS, against 11.2 and 12.2 on the real pairs. Most flagged pixels come from one scene, which holds 91\% of them on vanilla 3DGS. Without its three highest runs, vanilla's mean falls to 0.172\%. The figures for \gsdiff{} and PlenoCI are as PlenoCI reports them. Neither method's code is released, and PlenoCI does not state the reconstructions, views or number of pairs behind them, so they are listed beside ours but are not a measured comparison.

\section{CL-Splats}
\label{sec:clsplats}

CL-Splats~\cite{ackermann2025clsplats} photographs five real scenes twice, day 0 and day 1, with objects changed between them. It evaluates these scenes by rendering quality
alone and releases change masks only for its synthetic scenes, so we annotated
its evaluation views with PASLCD's convention.
The masks, the splits, the annotation protocol and the re-centred room camera below are ours, released with the paper~\cite{gspoolclsplats}. \gspool{} runs on these scenes with the configuration of the main experiments.

\textbf{Frame and split.} CL-Splats solves both days in one COLMAP model per scene, which corresponds to PASLCD's joint frame. Day 0 is the reference visit and day 1 the inference visit. Every eighth registered image by name, from the first, is held out. The held-out day-1 images are the 29 evaluation views, and all others train the reconstructions. The room's camera, whose principal point sits 3\,px from the
image centre that the 3DGS trainers assume, is re-centred with poses and points
unchanged.

\textbf{Masks.} As in PASLCD, an added object is marked where it appears and a
removed object where it stood, minus any part now hidden behind an unchanged object. Lighting and cast shadows are not marked. An added object's mask is
SAM2's~\cite{sam2} on the day-1 view. A removed object's region comes from
geometry: SAM2 tracks it through every day-0 photograph, the masks are carved
into a visual hull in the joint solve, and the hull is projected into each day-1
view, less the pixels where an unchanged object stands in front of it. When every sixth day-0 photograph is left out of the carving and the hull is projected into those photographs, it reproduces SAM2's masks at median IoU 0.93, 0.96 and 0.96 for the cone, bottle
and sneaker, and 0.81 for the thin folded card. Every mask was reviewed against
the photographs of both days, and no annotation setting was chosen against any
method's output.

\begin{figure*}[p]
\centering
\begin{minipage}[b]{0.6\textwidth}\centering\scriptsize\setlength{\tabcolsep}{3pt}
\begin{tabular}{@{}lcccccccc@{}}
\toprule
& \multicolumn{2}{c}{\gspool{} (FastPGSR)} & \multicolumn{2}{c}{\gspool{} (vanilla 3DGS)} & \multicolumn{2}{c}{\mvcd{}} & \multicolumn{2}{c}{\oscd{}} \\
\cmidrule(lr){2-3} \cmidrule(lr){4-5} \cmidrule(lr){6-7} \cmidrule(lr){8-9}
scene & mIoU & F1 & mIoU & F1 & mIoU & F1 & mIoU & F1 \\
\midrule
cone & 0.893 {\scriptsize[0.89--0.90]} & 0.943 & 0.818 {\scriptsize[0.81--0.82]} & 0.899 & 0.746 & 0.853 & 0.754 & 0.860 \\
cupboard & 0.647 {\scriptsize[0.55--0.79]} & 0.751 & 0.778 {\scriptsize[0.68--0.93]} & 0.864 & 0.539 & 0.687 & 0.469 & 0.616 \\
desk & 0.929 {\scriptsize[0.91--0.94]} & 0.962 & 0.948 {\scriptsize[0.94--0.95]} & 0.973 & 0.851 & 0.918 & \multicolumn{2}{c}{fails$^\dagger$} \\
room & 0.905 {\scriptsize[0.90--0.91]} & 0.949 & 0.916 {\scriptsize[0.92--0.92]} & 0.955 & 0.488 & 0.654 & 0.736 & 0.845 \\
shelf & 0.876 {\scriptsize[0.86--0.90]} & 0.929 & 0.909 {\scriptsize[0.90--0.91]} & 0.950 & 0.572 & 0.705 & 0.402 & 0.535 \\
\midrule
all 29 views & \textbf{0.8550} & \textbf{0.9114} & \textbf{0.8649} & \textbf{0.9237} & \textbf{0.6413} & \textbf{0.7667} & --- & --- \\
\bottomrule
\end{tabular}

\captionsetup{type=table}
\caption{Every CL-Splats scene on our annotations: \gspool{}'s seed-mean mIoU with the seed range in brackets and seed-mean F1, beside \mvcd{}'s and \oscd{}'s single runs ($^\dagger$no desk result).}
\label{tab:clsplatsperscene}
\end{minipage}\hfill
\begin{minipage}[b]{0.37\textwidth}\centering
\includegraphics[width=\linewidth]{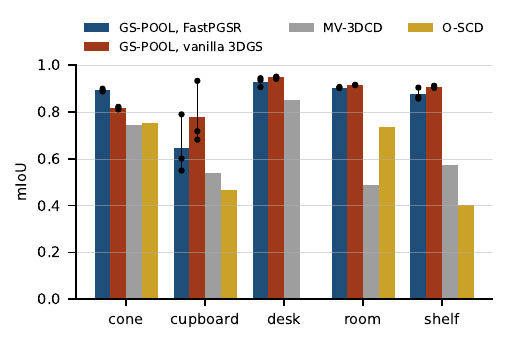}
\captionsetup{type=figure}
\caption{mIoU per CL-Splats scene: \gspool{}'s bar at the seed mean, whiskers at the seed range and dots at the seeds, and \mvcd{}'s and \oscd{}'s single runs (\oscd{} has no desk result).}
\label{fig:clsplatsperscene}
\end{minipage}

\vspace{1.5em}
\includegraphics[width=\textwidth]{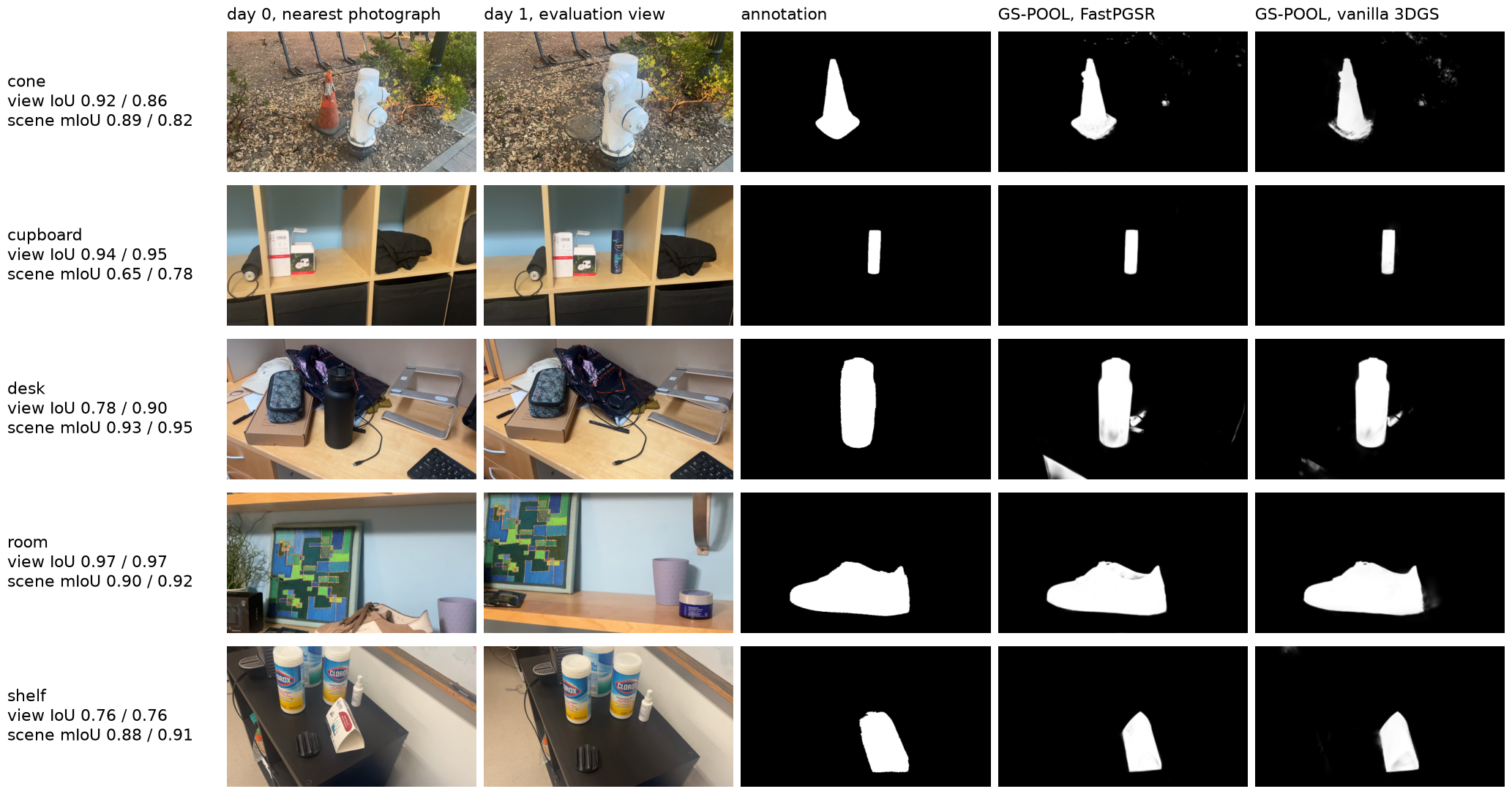}
\caption{All five CL-Splats scenes at the view with the most annotated change, both trainers at seed 22, labelled with the view's IoU and the scene's three-seed mIoU (FastPGSR / vanilla 3DGS).}
\label{fig:clsplatsall}
\end{figure*}

\textbf{Results.} Table~\ref{tab:clsplats} gives mIoU and F1 over the 29
evaluation views, for \gspool{} the mean over three seeds. \mvcd{} runs from its
released code at its defaults except its downscale: 2 rather than 8, which gives CL-Splats' 960\,px photographs the working resolution its recipe reaches on PASLCD. It is trained on the day-0 images and evaluated on every day-1 image. \oscd{} runs from its released code on a vanilla 3DGS reference of the day-0 training images and localises each day-1 image itself. It finds no pose for one desk image, so \oscd{} is scored on the other four scenes (25 views).
PlenoCI's figures, and those it reports for \gsdiff{}, \oscd{} and \mvcd{}, are on its own annotations and its split of every fifth image, with PlenoCI's own thresholds chosen per scene~\cite{plenoci}, so they are listed beside ours but not measured on the same masks. On our masks \mvcd{} scores within 1.2\% of what PlenoCI reports for it. Table~\ref{tab:clsplatsperscene} and Figure~\ref{fig:clsplatsperscene}
break the result into scenes, and Figure~\ref{fig:clsplatsall} shows all five.

\begin{table}[H]\centering\footnotesize
\caption{Change detection on the CL-Splats real scenes, PlenoCI's figures on its own annotations~\cite{plenoci} above and ours below ($^\dagger$four scenes, 25 views).}
\label{tab:clsplats}
\begin{tabular}{@{}llcc@{}}
\toprule
Method & annotations & mIoU & F1 \\
\midrule
\gsdiff{}~\cite{gsdiff} & PlenoCI & 0.724 & 0.829 \\
\oscd{}~\cite{oscd} & PlenoCI & 0.756 & 0.845 \\
\mvcd{}~\cite{mv3cd} & PlenoCI & 0.634 & 0.752 \\
PlenoCI~\cite{plenoci} & PlenoCI & 0.951 & 0.975 \\
\midrule
\mvcd{}~\cite{mv3cd} & \gspool{} & 0.6413 & 0.7667 \\
\oscd{}~\cite{oscd}$^\dagger$ & \gspool{} & 0.6224 & 0.7428 \\
\gspool{} (FastPGSR) & \gspool{} & 0.8550 & 0.9114 \\
\gspool{} (vanilla 3DGS) & \gspool{} & 0.8649 & 0.9237 \\
\bottomrule
\end{tabular}

\end{table}

\clearpage
\onecolumn
\section{Per-cell results}
\label{sec:allcells}\label{sec:percell}
\begin{multicols}{2}
\textbf{Scene-level uncertainty.} PASLCD has ten physical scenes, so a mean over them can rest on a few of them. We resample the ten scenes with replacement 100{,}000 times, keeping both instances and all three seeds within each block, and report percentile 95\% intervals, with paired differences on the same samples. On the joint frame, FastPGSR's mIoU interval is $[0.6825,0.8102]$, whose lower end is above \gsdiff{}'s reported 0.644, and vanilla 3DGS's is $[0.6371,0.7972]$. FastPGSR leads vanilla by 4.3\%, with interval $[0.2\%, 9.1\%]$. The intervals describe these ten scenes under a fixed configuration.

\begin{figure*}[p]\centering
\includegraphics[width=0.92\textwidth]{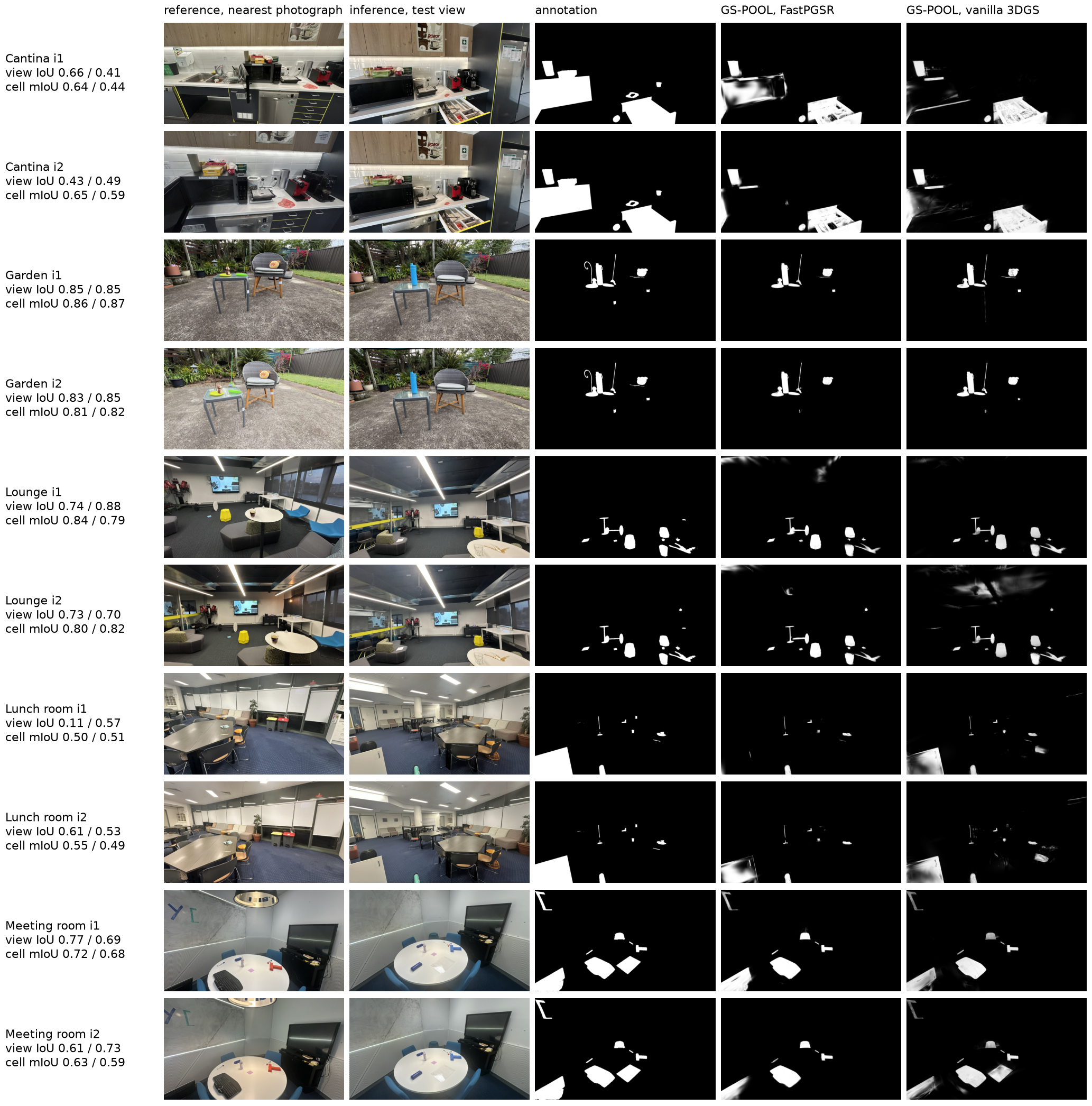}
\caption{Cells 1--10 (Cantina to Meeting\_room) at the test view with the most annotated change, joint frame, seed 22, labelled with the view's IoU and the cell's three-seed mIoU (FastPGSR / vanilla 3DGS).}
\label{fig:allcells1}
\end{figure*}

\begin{figure*}[p]\centering
\includegraphics[width=0.92\textwidth]{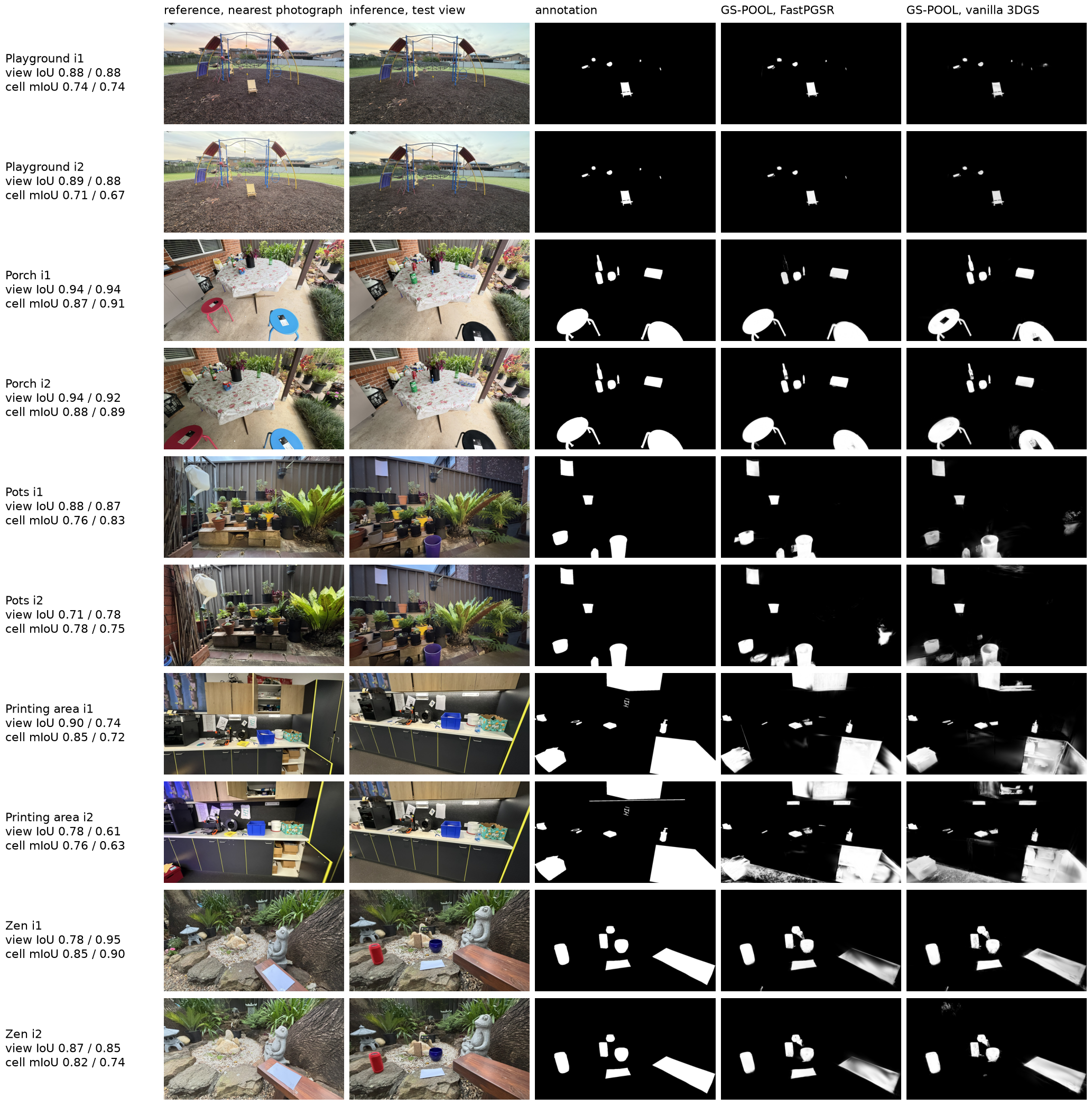}
\caption{Cells 11--20 (Playground to Zen), as in Figure~\ref{fig:allcells1}.}
\label{fig:allcells2}
\end{figure*}

\textbf{Cells.} Table~\ref{tab:percell} and Figure~\ref{fig:percell} break
Table~\ref{tab:main} into its 20 cells, and Figures~\ref{fig:allcells1} and~\ref{fig:allcells2} show every cell at its test view with the largest annotated change, chosen by the size of the change and not by its score.
\end{multicols}
\begin{center}\captionsetup{type=table}
\begingroup\centering\footnotesize\setlength{\tabcolsep}{4pt}
\begin{tabular}{@{}lcccccccc@{}}
\toprule
& \multicolumn{2}{c}{FastPGSR, joint} & \multicolumn{2}{c}{FastPGSR, anchored} & \multicolumn{2}{c}{vanilla 3DGS, joint} & \multicolumn{2}{c}{vanilla 3DGS, anchored} \\
\cmidrule(lr){2-3} \cmidrule(lr){4-5} \cmidrule(lr){6-7} \cmidrule(lr){8-9}
cell & mIoU & F1 & mIoU & F1 & mIoU & F1 & mIoU & F1 \\
\midrule
Cantina\_i1 & 0.642 {\scriptsize[0.63--0.65]} & 0.773 & 0.621 {\scriptsize[0.60--0.67]} & 0.758 & 0.437 {\scriptsize[0.42--0.46]} & 0.600 & 0.545 {\scriptsize[0.40--0.62]} & 0.684 \\
Cantina\_i2 & 0.647 {\scriptsize[0.61--0.67]} & 0.779 & 0.618 {\scriptsize[0.59--0.65]} & 0.758 & 0.593 {\scriptsize[0.56--0.63]} & 0.736 & 0.588 {\scriptsize[0.58--0.59]} & 0.730 \\
Garden\_i1 & 0.855 {\scriptsize[0.85--0.86]} & 0.922 & 0.848 {\scriptsize[0.84--0.85]} & 0.917 & 0.870 {\scriptsize[0.86--0.88]} & 0.930 & 0.858 {\scriptsize[0.86--0.86]} & 0.924 \\
Garden\_i2 & 0.807 {\scriptsize[0.80--0.81]} & 0.893 & 0.607 {\scriptsize[0.45--0.75]} & 0.747 & 0.824 {\scriptsize[0.81--0.83]} & 0.903 & 0.749 {\scriptsize[0.74--0.75]} & 0.855 \\
Lounge\_i1 & 0.839 {\scriptsize[0.82--0.85]} & 0.910 & 0.818 {\scriptsize[0.78--0.84]} & 0.895 & 0.789 {\scriptsize[0.68--0.85]} & 0.874 & 0.829 {\scriptsize[0.79--0.86]} & 0.903 \\
Lounge\_i2 & 0.805 {\scriptsize[0.78--0.85]} & 0.890 & 0.791 {\scriptsize[0.76--0.84]} & 0.881 & 0.818 {\scriptsize[0.75--0.86]} & 0.897 & 0.807 {\scriptsize[0.74--0.85]} & 0.891 \\
Lunch\_room\_i1 & 0.498 {\scriptsize[0.45--0.59]} & 0.634 & 0.521 {\scriptsize[0.49--0.58]} & 0.657 & 0.515 {\scriptsize[0.47--0.57]} & 0.652 & 0.517 {\scriptsize[0.46--0.59]} & 0.650 \\
Lunch\_room\_i2 & 0.547 {\scriptsize[0.50--0.59]} & 0.682 & 0.489 {\scriptsize[0.47--0.52]} & 0.635 & 0.495 {\scriptsize[0.47--0.54]} & 0.632 & 0.574 {\scriptsize[0.57--0.58]} & 0.692 \\
Meeting\_room\_i1 & 0.724 {\scriptsize[0.71--0.74]} & 0.838 & 0.722 {\scriptsize[0.72--0.73]} & 0.837 & 0.678 {\scriptsize[0.65--0.69]} & 0.805 & 0.731 {\scriptsize[0.68--0.83]} & 0.841 \\
Meeting\_room\_i2 & 0.630 {\scriptsize[0.63--0.63]} & 0.767 & 0.565 {\scriptsize[0.56--0.57]} & 0.711 & 0.587 {\scriptsize[0.50--0.75]} & 0.724 & 0.513 {\scriptsize[0.43--0.59]} & 0.666 \\
Playground\_i1 & 0.738 {\scriptsize[0.73--0.74]} & 0.837 & 0.737 {\scriptsize[0.73--0.74]} & 0.836 & 0.743 {\scriptsize[0.74--0.75]} & 0.840 & 0.752 {\scriptsize[0.74--0.77]} & 0.848 \\
Playground\_i2 & 0.714 {\scriptsize[0.71--0.72]} & 0.821 & 0.711 {\scriptsize[0.71--0.72]} & 0.818 & 0.669 {\scriptsize[0.66--0.68]} & 0.786 & 0.681 {\scriptsize[0.67--0.69]} & 0.795 \\
Porch\_i1 & 0.871 {\scriptsize[0.87--0.87]} & 0.929 & 0.864 {\scriptsize[0.85--0.88]} & 0.925 & 0.911 {\scriptsize[0.91--0.91]} & 0.953 & 0.902 {\scriptsize[0.90--0.91]} & 0.948 \\
Porch\_i2 & 0.882 {\scriptsize[0.87--0.90]} & 0.936 & 0.888 {\scriptsize[0.88--0.89]} & 0.940 & 0.887 {\scriptsize[0.88--0.90]} & 0.939 & 0.882 {\scriptsize[0.88--0.88]} & 0.937 \\
Pots\_i1 & 0.758 {\scriptsize[0.70--0.82]} & 0.859 & 0.822 {\scriptsize[0.82--0.83]} & 0.902 & 0.827 {\scriptsize[0.81--0.85]} & 0.905 & 0.789 {\scriptsize[0.75--0.82]} & 0.880 \\
Pots\_i2 & 0.780 {\scriptsize[0.76--0.81]} & 0.874 & 0.721 {\scriptsize[0.66--0.76]} & 0.835 & 0.748 {\scriptsize[0.73--0.76]} & 0.854 & 0.556 {\scriptsize[0.53--0.58]} & 0.712 \\
Printing\_area\_i1 & 0.853 {\scriptsize[0.84--0.86]} & 0.919 & 0.826 {\scriptsize[0.81--0.84]} & 0.903 & 0.722 {\scriptsize[0.70--0.76]} & 0.834 & 0.745 {\scriptsize[0.73--0.77]} & 0.847 \\
Printing\_area\_i2 & 0.755 {\scriptsize[0.65--0.81]} & 0.854 & 0.612 {\scriptsize[0.60--0.63]} & 0.746 & 0.635 {\scriptsize[0.56--0.74]} & 0.761 & 0.722 {\scriptsize[0.72--0.72]} & 0.831 \\
Zen\_i1 & 0.852 {\scriptsize[0.84--0.87]} & 0.917 & 0.875 {\scriptsize[0.84--0.92]} & 0.931 & 0.905 {\scriptsize[0.87--0.94]} & 0.944 & 0.918 {\scriptsize[0.88--0.94]} & 0.955 \\
Zen\_i2 & 0.819 {\scriptsize[0.81--0.83]} & 0.886 & 0.864 {\scriptsize[0.82--0.89]} & 0.923 & 0.742 {\scriptsize[0.74--0.75]} & 0.849 & 0.904 {\scriptsize[0.88--0.93]} & 0.946 \\
\midrule
mean & \textbf{0.7509} & \textbf{0.8461} & \textbf{0.7260} & \textbf{0.8277} & \textbf{0.7198} & \textbf{0.8209} & \textbf{0.7281} & \textbf{0.8266} \\
\bottomrule
\end{tabular}
\caption{Every PASLCD cell under the four configurations: seed-mean mIoU with the seed range in brackets and seed-mean F1.}
\label{tab:percell}
\endgroup

\vspace{0.8em}
\includegraphics[width=0.72\textwidth]{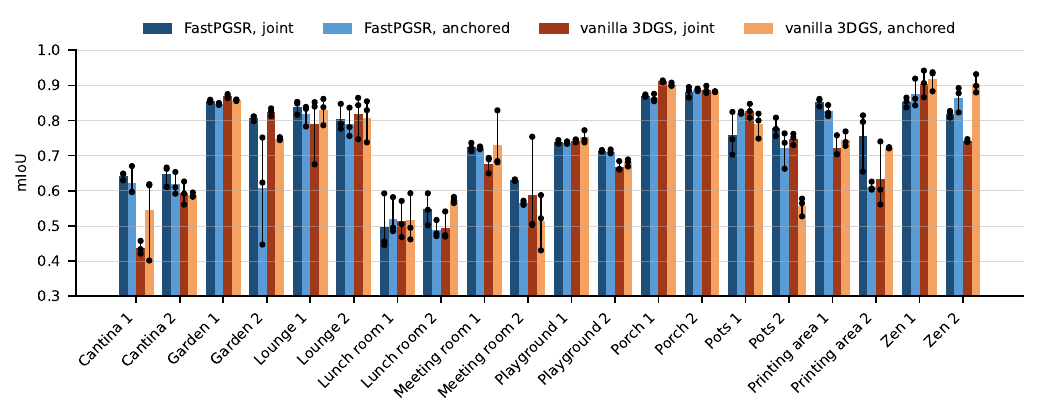}
\captionof{figure}{mIoU per cell for the four configurations: bar at the seed mean, whiskers at the seed range, dots at the seeds.}
\label{fig:percell}
\end{center}
\clearpage



\begin{thebibliography}{99}\small

\bibitem{3dgs} B.~Kerbl, G.~Kopanas, T.~Leimk\"uhler, and G.~Drettakis. 3D Gaussian Splatting for real-time radiance field rendering. \emph{ACM Transactions on Graphics}, 42(4), 2023.

\bibitem{paslcd} C.~J. Galappaththige, J.~Lai, L.~Windrim, D.~Dansereau, N.~S\"underhauf, and D.~Miller. PASLCD: Pose-Agnostic Scene-Level Change Detection dataset. \url{https://huggingface.co/datasets/ChamudithaJay/PASLCD}, 2025.

\bibitem{plenoci} J.~Lai, C.~J. Galappaththige, N.~S\"underhauf, D.~Miller, and D.~G. Dansereau. PlenoCI: Plenoptic CharacterIstics for view dependence aware change classification. arXiv:2609.28930, 2026.

\bibitem{gsdiff} C.~J. Galappaththige, J.~Lai, T.~Patten, D.~Dansereau, N.~S\"underhauf, and D.~Miller. From pixels to primitives: Scene change detection in 3D Gaussian Splatting. arXiv:2605.07203, 2026.

\bibitem{ackermann2025clsplats} J.~Ackermann, J.~Kulhanek, S.~Cai, H.~Xu, M.~Pollefeys, G.~Wetzstein, L.~Guibas, and S.~Peng. CL-Splats: Continual Learning of Gaussian Splatting with Local Optimization. In \emph{IEEE/CVF International Conference on Computer Vision}, 2025.

\bibitem{sakurada_change_2015} K.~Sakurada and T.~Okatani. Change detection from a street image pair using CNN features and superpixel segmentation. In \emph{British Machine Vision Conference}, 2015.

\bibitem{alcantarilla2018street} P.~F. Alcantarilla, S.~Stent, G.~Ros, R.~Arroyo, and R.~Gherardi. Street-view change detection with deconvolutional networks. \emph{Autonomous Robots}, 42(7):1301--1322, 2018.

\bibitem{daudt2018fully} R.~C. Daudt, B.~Le Saux, and A.~Boulch. Fully convolutional siamese networks for change detection. In \emph{IEEE International Conference on Image Processing}, 2018.

\bibitem{varghese2018changenet} A.~Varghese, J.~Gubbi, A.~Ramaswamy, and P.~Balamuralidhar. ChangeNet: A deep learning architecture for visual change detection. In \emph{European Conference on Computer Vision Workshops}, 2018.

\bibitem{cyws2d} R.~Sachdeva and A.~Zisserman. The change you want to see. In \emph{IEEE/CVF Winter Conference on Applications of Computer Vision}, 2023.

\bibitem{cyws3d} R.~Sachdeva and A.~Zisserman. The change you want to see (now in 3D). In \emph{IEEE/CVF International Conference on Computer Vision Workshops}, 2023.

\bibitem{lin2025robust} C.-J.~Lin, S.~Garg, T.-J.~Chin, and F.~Dayoub. Robust scene change detection using visual foundation models and cross-attention mechanisms. In \emph{IEEE International Conference on Robotics and Automation}, 2025.

\bibitem{kim2025towards} J.-W.~Kim and U.-H.~Kim. Towards Generalizable Scene Change Detection. In \emph{IEEE/CVF Conference on Computer Vision and Pattern Recognition}, 2025.

\bibitem{cho2025zero} K.~Cho, D.~Y. Kim, and E.~Kim. Zero-shot scene change detection. In \emph{AAAI Conference on Artificial Intelligence}, 2025.

\bibitem{goldilocs} A.~Friedlander, A.~Shamir, and O.~Fried. GOLDILOCS: General Object-Level Detection and Labeling of Changes in Scenes. In \emph{International Conference on Learning Representations}, 2026.

\bibitem{mv3cd} C.~J. Galappaththige, J.~Lai, L.~Windrim, D.~G. Dansereau, N.~S\"underhauf, and D.~Miller. Multi-View Pose-Agnostic Change Localization with Zero Labels. In \emph{IEEE/CVF Conference on Computer Vision and Pattern Recognition}, 2025.

\bibitem{kruse2024splatpose} M.~Kruse, M.~Rudolph, D.~Woiwode, and B.~Rosenhahn. SplatPose \& Detect: Pose-agnostic 3D anomaly detection. In \emph{IEEE/CVF Conference on Computer Vision and Pattern Recognition Workshops}, 2024.

\bibitem{liu2024splatposeplus} Y.~Liu, Y.~S. Hu, Y.~Chen, and J.~Zelek. SplatPose+: Real-Time Image-Based Pose-Agnostic 3D Anomaly Detection. In \emph{European Conference on Computer Vision Workshops}, 2024.

\bibitem{zhou2023pad} Q.~Zhou, W.~Li, L.~Jiang, G.~Wang, G.~Zhou, S.~Zhang, and H.~Zhao. PAD: A dataset and benchmark for pose-agnostic anomaly detection. In \emph{Advances in Neural Information Processing Systems}, 2023.

\bibitem{lu20253dgs} Z.~Lu, J.~Ye, and J.~Leonard. 3DGS-CD: 3D Gaussian splatting-based change detection for physical object rearrangement. \emph{IEEE Robotics and Automation Letters}, 10(3):2662--2669, 2025.

\bibitem{jiang2025gaussian} B.~Jiang, R.~Huang, Q.~Zhao, and Y.~Zhang. Gaussian Difference: Find any change instance in 3D scenes. In \emph{IEEE International Conference on Acoustics, Speech, and Signal Processing}, 2025.

\bibitem{oscd} C.~J. Galappaththige, J.~Lai, L.~Windrim, D.~G. Dansereau, N.~S\"underhauf, and D.~Miller. Changes in Real Time: Online Scene Change Detection with Multi-View Fusion. In \emph{IEEE/CVF Conference on Computer Vision and Pattern Recognition}, 2026.

\bibitem{u3dgs} C.~J. Galappaththige, T.~Gottwald, P.~Stehr, E.~Heinert, N.~S\"underhauf, D.~Miller, and M.~Rottmann. Predictive photometric uncertainty in Gaussian Splatting for novel view synthesis. In \emph{European Conference on Computer Vision}, 2026.

\bibitem{scenediff} Y.~Wu, C.~Lin, H.~Che, A.~Tiwari, C.~Zou, S.~Wang, and D.~Hoiem. SceneDiff: A benchmark and method for multiview object change detection. In \emph{European Conference on Computer Vision}, pages 637--656, 2026.

\bibitem{c2c} D.~Girardeau-Montaut, M.~Roux, R.~Marc, and G.~Thibault. Change detection on points cloud data acquired with a ground laser scanner. \emph{International Archives of Photogrammetry, Remote Sensing and Spatial Information Sciences}, XXXVI-3/W19:30--35, 2005.

\bibitem{LAGUE201310} D.~Lague, N.~Brodu, and J.~Leroux. Accurate 3D comparison of complex topography with terrestrial laser scanner: Application to the Rangitikei canyon (N-Z). \emph{ISPRS Journal of Photogrammetry and Remote Sensing}, 82:10--26, 2013.

\bibitem{yew2021city} Z.~J. Yew and G.~H. Lee. City-scale scene change detection using point clouds. In \emph{IEEE International Conference on Robotics and Automation}, 2021.

\bibitem{de2023siamese} I.~de G{\'e}lis, S.~Lef{\`e}vre, and T.~Corpetti. Siamese KPConv: 3D multiple change detection from raw point clouds using deep learning. \emph{ISPRS Journal of Photogrammetry and Remote Sensing}, 197:274--291, 2023.

\bibitem{schmid2022panoptic} L.~Schmid, J.~Delmerico, J.~L. Sch{\"o}nberger, J.~Nieto, M.~Pollefeys, R.~Siegwart, and C.~Cadena. Panoptic Multi-TSDFs: a flexible representation for online multi-resolution volumetric mapping and long-term dynamic scene consistency. In \emph{IEEE International Conference on Robotics and Automation}, 2022.

\bibitem{fu2022planesdf} J.~Fu, C.~Lin, Y.~Taguchi, A.~Cohen, Y.~Zhang, S.~Mylabathula, and J.~J. Leonard. PlaneSDF-based change detection for long-term dense mapping. \emph{IEEE Robotics and Automation Letters}, 7(4):9667--9674, 2022.

\bibitem{zhu2024living} L.~Zhu, S.~Huang, K.~Schindler, and I.~Armeni. Living scenes: Multi-object relocalization and reconstruction in changing 3D environments. In \emph{IEEE/CVF Conference on Computer Vision and Pattern Recognition}, 2024.

\bibitem{taneja2011image} A.~Taneja, L.~Ballan, and M.~Pollefeys. Image based detection of geometric changes in urban environments. In \emph{IEEE International Conference on Computer Vision}, 2011.

\bibitem{cheng2025lt} L.~Cheng, Z.~Qi, Z.~Zhou, C.~Lu, and G.~Xiong. LT-Gaussian: Long-Term Map Update Using 3D Gaussian Splatting for Autonomous Driving. In \emph{IEEE Intelligent Vehicles Symposium}, 2025.

\bibitem{zeng2025gaussianupdate} L.~Zeng, B.~Zhao, J.~Hu, X.~Shen, Z.~Dang, H.~Bao, and Z.~Cui. GaussianUpdate: Continual 3D Gaussian Splatting Update for Changing Environments. In \emph{IEEE/CVF International Conference on Computer Vision}, 2025.

\bibitem{yugay2025gaussian} V.~Yugay, T.~Kersten, L.~Carlone, T.~Gevers, M.~R. Oswald, and L.~Schmid. Gaussian mapping for evolving scenes. In \emph{IEEE/CVF Conference on Computer Vision and Pattern Recognition}, pages 18903--18912, 2026.

\bibitem{gaussreg} J.~Chang, Y.~Xu, Y.~Li, Y.~Chen, W.~Feng, and X.~Han. GaussReg: Fast 3D registration with Gaussian splatting. In \emph{European Conference on Computer Vision}, 2024.

\bibitem{siren} O.~Shorinwa, J.~Sun, M.~Schwager, and A.~Majumdar. SIREN: Semantic, initialization-free registration of multi-robot Gaussian splatting maps. In \emph{Conference on Robot Learning}, 2025. arXiv:2502.06519.

\bibitem{dff} S.~Kobayashi, E.~Matsumoto, and V.~Sitzmann. Decomposing NeRF for editing via feature field distillation. In \emph{Advances in Neural Information Processing Systems}, 2022.

\bibitem{n3f} V.~Tschernezki, I.~Laina, D.~Larlus, and A.~Vedaldi. Neural Feature Fusion Fields: 3D distillation of self-supervised 2D image representations. In \emph{International Conference on 3D Vision}, 2022.

\bibitem{lerf} J.~Kerr, C.~M. Kim, K.~Goldberg, A.~Kanazawa, and M.~Tancik. LERF: Language embedded radiance fields. In \emph{IEEE/CVF International Conference on Computer Vision}, 2023.

\bibitem{feat3dgs} S.~Zhou, H.~Chang, S.~Jiang, Z.~Fan, Z.~Zhu, D.~Xu, P.~Chari, S.~You, Z.~Wang, and A.~Kadambi. Feature 3DGS: Supercharging 3D Gaussian splatting to enable distilled feature fields. In \emph{IEEE/CVF Conference on Computer Vision and Pattern Recognition}, 2024.

\bibitem{langsplat} M.~Qin, W.~Li, J.~Zhou, H.~Wang, and H.~Pfister. LangSplat: 3D language Gaussian splatting. In \emph{IEEE/CVF Conference on Computer Vision and Pattern Recognition}, 2024.

\bibitem{gaussiangrouping} M.~Ye, M.~Danelljan, F.~Yu, and L.~Ke. Gaussian Grouping: Segment and edit anything in 3D scenes. In \emph{European Conference on Computer Vision}, 2024.

\bibitem{dinov3} O.~Sim\'eoni, H.~V. Vo, M.~Seitzer, F.~Baldassarre, M.~Oquab, C.~Jose, V.~Khalidov, M.~Szafraniec, S.~Yi, M.~Ramamonjisoa, F.~Massa, D.~Haziza, L.~Wehrstedt, J.~Wang, T.~Darcet, T.~Moutakanni, L.~Sentana, C.~Roberts, A.~Vedaldi, J.~Tolan, J.~Brandt, C.~Couprie, J.~Mairal, H.~J\'egou, P.~Labatut, and P.~Bojanowski. DINOv3. arXiv:2508.10104, 2025.

\bibitem{sam2} N.~Ravi, V.~Gabeur, Y.-T. Hu, R.~Hu, C.~Ryali, T.~Ma, H.~Khedr, R.~R\"adle, C.~Rolland, L.~Gustafson, E.~Mintun, J.~Pan, K.~V. Alwala, N.~Carion, C.-Y. Wu, R.~Girshick, P.~Doll\'ar, and C.~Feichtenhofer. SAM 2: Segment anything in images and videos. In \emph{International Conference on Learning Representations}, 2025.

\bibitem{fastgs} S.~Ren, T.~Wen, Y.~Fang, and B.~Lu. FastGS: Training 3D Gaussian Splatting in 100 Seconds. In \emph{IEEE/CVF Conference on Computer Vision and Pattern Recognition}, 2026.

\bibitem{pgsr} D.~Chen, H.~Li, W.~Ye, Y.~Wang, W.~Xie, S.~Zhai, N.~Wang, H.~Liu, H.~Bao, and G.~Zhang. PGSR: Planar-Based Gaussian Splatting for efficient and high-fidelity surface reconstruction. \emph{IEEE Transactions on Visualization and Computer Graphics}, 31(9):6100--6111, 2025.

\bibitem{gsplat} V.~Ye, R.~Li, J.~Kerr, M.~Turkulainen, B.~Yi, Z.~Pan, O.~Seiskari, J.~Ye, J.~Hu, M.~Tancik, and A.~Kanazawa. gsplat: An open-source library for Gaussian Splatting. \emph{Journal of Machine Learning Research}, 26(34):1--17, 2025.

\bibitem{flashsplat} Q.~Shen, X.~Yang, and X.~Wang. FlashSplat: 2D to 3D Gaussian Splatting segmentation solved optimally. In \emph{European Conference on Computer Vision}, 2024.

\bibitem{ssim} Z.~Wang, A.~C. Bovik, H.~R. Sheikh, and E.~P. Simoncelli. Image quality assessment: From error visibility to structural similarity. \emph{IEEE Transactions on Image Processing}, 13(4):600--612, 2004.

\bibitem{mannwhitney} H.~B. Mann and D.~R. Whitney. On a test of whether one of two random variables is stochastically larger than the other. \emph{Annals of Mathematical Statistics}, 18(1):50--60, 1947.

\bibitem{fisher} R.~A. Fisher. \emph{Statistical Methods for Research Workers}, 4th ed. Oliver and Boyd, Edinburgh, 1932.

\bibitem{brown} M.~B. Brown. A method for combining non-independent, one-sided tests of significance. \emph{Biometrics}, 31(4):987--992, 1975.

\bibitem{ebm} W.~Poole, D.~L. Gibbs, I.~Shmulevich, B.~Bernard, and T.~A. Knijnenburg. Combining dependent P-values with an empirical adaptation of Brown's method. \emph{Bioinformatics}, 32(17):i430--i436, 2016.

\bibitem{mahalanobis} P.~C. Mahalanobis. On the generalised distance in statistics. \emph{Proceedings of the National Institute of Sciences of India}, 2(1):49--55, 1936.

\bibitem{jiang2024fisherrf} W.~Jiang, B.~Lei, and K.~Daniilidis. FisherRF: Active view selection and mapping with radiance fields using Fisher information. In \emph{European Conference on Computer Vision}, 2024.

\bibitem{colmap} J.~L. Sch\"onberger and J.-M. Frahm. Structure-from-motion revisited. In \emph{IEEE Conference on Computer Vision and Pattern Recognition}, 2016.

\bibitem{gspoolpaslcd} \iftoggle{cvprfinal}{B.~Keren-Gil, J.~Gain, and P.~Marais. \gspool{} on PASLCD: everything the paper measured. \url{https://huggingface.co/datasets/boazkg/gs-pool-paslcd}, 2026.}{Anonymous. \gspool{} on PASLCD: everything the paper measured. Link withheld for review.}

\bibitem{gspoolclsplats} \iftoggle{cvprfinal}{B.~Keren-Gil, J.~Gain, and P.~Marais. \gspool{} on CL-Splats: change masks for the real-world scenes. \url{https://huggingface.co/datasets/boazkg/gs-pool-cl-splats}, 2026.}{Anonymous. \gspool{} on CL-Splats: change masks for the real-world scenes. Link withheld for review.}
\end{thebibliography}
\end{document}